\documentclass[10pt, conference]{IEEEtran}
\usepackage[utf8]{inputenc}

\usepackage{amsmath,amssymb,amsfonts}
\usepackage{graphicx}
\usepackage{textcomp}
\usepackage{pifont}

\usepackage[detect-none]{siunitx}
\usepackage{amsthm}
\usepackage{xcolor}
\usepackage[figuresright]{rotating}

\usepackage{graphicx}
\graphicspath{{/}{fig/}}

\usepackage{array}
\usepackage{textcomp}
\usepackage{xcolor}
\usepackage{multirow}
\usepackage{booktabs}

\usepackage{mathtools}
\usepackage{breqn}
\usepackage{float}

\usepackage{pgfplots}
\pgfplotsset{compat=1.7}
\usepgfplotslibrary{groupplots}

\usepackage{multirow,tabularx}
\usepackage{hyperref}
\usepackage{flushend}

\newcommand{\yess}{\scriptsize{\ding{52}}}

\newlength\figureheight
\newlength\figurewidth
\title{\huge
    UWB-based system for UAV Localization in \\ GNSS-Denied Environments: Characterization and Dataset
}

\author{
    \IEEEauthorblockN{
        \vspace{1em}
        Jorge Peña Queralta\IEEEauthorrefmark{1},
        Carmen Martínez Almansa\IEEEauthorrefmark{1},
        Fabrizio Schiano\IEEEauthorrefmark{2},
        Dario Floreano\IEEEauthorrefmark{2},
        Tomi Westerlund\IEEEauthorrefmark{1}
    }\\[+1em]
    \IEEEauthorblockA{
        \normalsize
        \IEEEauthorrefmark{1}\href{https://tiers.utu.fi}{Turku Intelligent Embedded and Robotic Systems (TIERS) Lab, University of Turku, Turku, Finland}.\\
        \IEEEauthorrefmark{2}\href{https://www.epfl.ch/labs/lis/}{Laboratory of Intelligent Systems (LIS)}, École Polytechnique Fédérale de Lausanne (EPFL), 1015 Lausanne, Switzerland.\\
        Emails: \IEEEauthorrefmark{1}\{jopequ, camart, tovewe\}@utu.fi, \IEEEauthorrefmark{2}\{fabrizio.schiano, dario.floreano\}@epfl.ch
    }
}

\begin{document}

\maketitle
\thispagestyle{empty}
\pagestyle{empty}

\begin{abstract}

    Small unmanned aerial vehicles (UAV) have penetrated multiple domains over the past years. In GNSS-denied or indoor environments, aerial robots require a robust and stable localization system, often with external feedback, in order to fly safely. Motion capture systems are typically utilized indoors when accurate localization is needed. However, these systems are expensive and most require a fixed setup. 
    %
    In this paper, we study and characterize an ultra-wideband (UWB) system for navigation and localization of aerial robots indoors based on Decawave's DWM1001 UWB node. The system is portable, inexpensive and can be battery powered in its totality. We show the viability of this system for autonomous flight of UAVs, and provide open-source methods and data that enable its widespread application even with movable anchor systems. We characterize the accuracy based on the position of the UAV with respect to the anchors, its altitude and speed, and the distribution of the anchors in space. Finally, we analyze the accuracy of the self-calibration of the anchors' positions.
    %
    %

\end{abstract}

\begin{IEEEkeywords}
    Robotics; UWB; UAV; Localization; Aerial Robotics; Unmanned Aerial Vehicles; UWB Positioning; 
\end{IEEEkeywords}

\IEEEpeerreviewmaketitle


\section{Introduction}

Autonomous unmanned aerial vehicles (UAVs) have gained increasing traction over the past years. The autonomous operation of these vehicles outdoors is mainly based on GNSS sensors~\cite{alsalam2017autonomous}. Nonetheless, many applications require the robots to operate in GNSS-denied environments, from factories or warehouses~\cite{tiemann2017scalable} to post-disaster and emergency scenarios~\cite{cui2015drones}. We study a wireless positioning system that requires active beacons both onboard the robots and in known positions in the operational environment. While the positions of a subset of the beacons, or anchors, must be known, we analyze the accuracy of their autopositioning. Having accurate localization has additional benefits in multi-robot systems, wither for collaborative mapping~\cite{queralta2019collaborative}, or pattern configuration~\cite{mccord2019progressive} purposes.

We investigate the properties of a mobile and inexpensive ultra-wideband (UWB) wireless positioning system that can be quickly deployed in GNSS-denied emergency scenarios, or in general for indoor environments. Compared with other indoor localization systems such as motion captures~\cite{furtado2019comparative}, a drastic decrease in both system complexity and price only has a relatively small impact on positioning accuracy. More importantly, even if the accuracy is reduced, the localization estimation is stable and does not threaten the smooth flight of a UAV. These type of system can complement existing motion capture (MOCAP) systems providing more flexible and mobile deployments.

Some commercial UAVs already utilize UWB for indoor localization. For instance, Bitcraze's Crazyflie~\cite{giernacki2017crazyflie}, with its Loco positioning deck, utilizes a UWB tag for indoor positioning. The Loco add-on relies on Decawave's DWM1000. In this paper, we work with the latest generation of UWB transceivers, the DWM1001, that provide improved localization accuracy. Another company that utilizes a similar, but undisclosed, technology, is Verity Studios~\cite{cirque2016zurich}. Verity develops multi-UAV systems for light shows indoors. In terms of UWB localization systems, higher-end solutions integrated within ready-to-use systems are available from vendors other than Decawave, such as Pozyx~\cite{pozyx2018pozyx}, Sewio and OpenRTLS~\cite{contigiani2016implementation}.

UWB wireless localization technologies have gained increasing attention in mobile robot applications in the past few years. UWB is a mature technology that has been studied for over two decades~\cite{fontana2000ultra}, with the IEEE 802.15.4a standard including specifications for UWB over a decade ago. UWB systems can be utilized for communication and localization~\cite{sahinoglu2008ultra}, or as short-range radar systems~\cite{taylor2018ultra}. UWB systems enable localization of a mobile tag from distance-only measurements between the tag and fixed anchor nodes with known position. The distance can be estimated via either time-of-flight (ToF) or time difference of arrival (TDoA). In the former case, the tag can calculate the distance to each of the anchors separately, while the latter estimation requires all anchors to be either connected in a local network or have a very accurate clock synchronization~\cite{mazraani2017experimental}.

More recently, with increased accuracy and more commercially available radios, UWB has been applied for indoor positioning and navigation in the field of mobile robotics~\cite{wang2017ultra, song2019uwb}. UWB-based positioning has been applied in mobile robots to aid navigation as part of multi-modal simultaneous localization and mapping (SLAM) algorithms~\cite{perez2017multi}, or for aiding odometry~\cite{magnago2019robot}. In the field of aerial robotics, it can aid vision sensors during the approximation for docking in a moving platform~\cite{nguyen2019integrated}, or for navigation in warehouses~\cite{macoir2019uwb}.


Several datasets and analysis reports exist for indoor localization of mobile robots based on UWB~\cite{raza2019dataset, barral2019dataset}, including UAVs~\cite{li2018dronedataset}. However, we have found that all previous studies involving the localization of UAVs are based on TDoA measurements which are more accurate but limit significantly the mobility of the system as a whole. Instead, we rely on ToF distance estimation only and analyze the self-calibration of anchor positions for a mobile setting. 
Finally, we study the localization accuracy as a function of the UAVs speed, height and position with respect to the anchors, both within and outside the convex envelope defined by the anchor positions. This type of characterization based on the operational details of the UAV does not exist in previous works. In this paper, we utilize Decawave DWM1001 UWB modules, the latest generation with an advertised accuracy of up to 5~cm. The data is acquired using the Robot Operating System (ROS). Custom ROS packages have been written for interfacing with the DWM1001 modules depending on their configuration (anchor, active tag, passive tag) and made publicly available, while the position information is given by Decawave's UART API, which is closed source.

The main contributions of this paper are the following. First, the introduction of a novel dataset that relies on ToF measurements of UWB signals for positioning of UAVs, meant for fast and mobile deployments with ground robots acting as anchors. This includes an accuracy and latency analysis of the auto-positioning of the anchors. Second, the characterization of the UAV positioning accuracy as a function of the spatial distribution of the anchors, the distance of the UAV to the center of mass of the anchors, and its speed and height. Moreover, initial experiments show the feasibility of stable autonomous flight based on the proposed localization system. We also provide open-source code for the automatic calibration of anchor positions, as well as the ROS nodes used for interfacing with the UWB devices in different modes.

The rest of this paper is organized as follows. Section II reviews the state-of-the-art in UWB localization for mobile robots, comparing our approach with existing datasets for UWB-based positioning in GNSS-denied environments. In Section III, we introduce our dataset, with ToF measurements and location estimations for different anchor configurations and varying number of mobile tags. Section IV then introduces initial experiments for UWB-based autonomous flight, together with a characterization of the localization accuracy for different anchor distributions (separation and height) in terms of the UAV state (position, height, speed). Finally, Section V concludes the work and outlines future research directions.




\begin{figure}
    \centering
    \begin{minipage}{0.24\textwidth}
        \centering
        \includegraphics[width=\textwidth, height=0.6\textwidth]{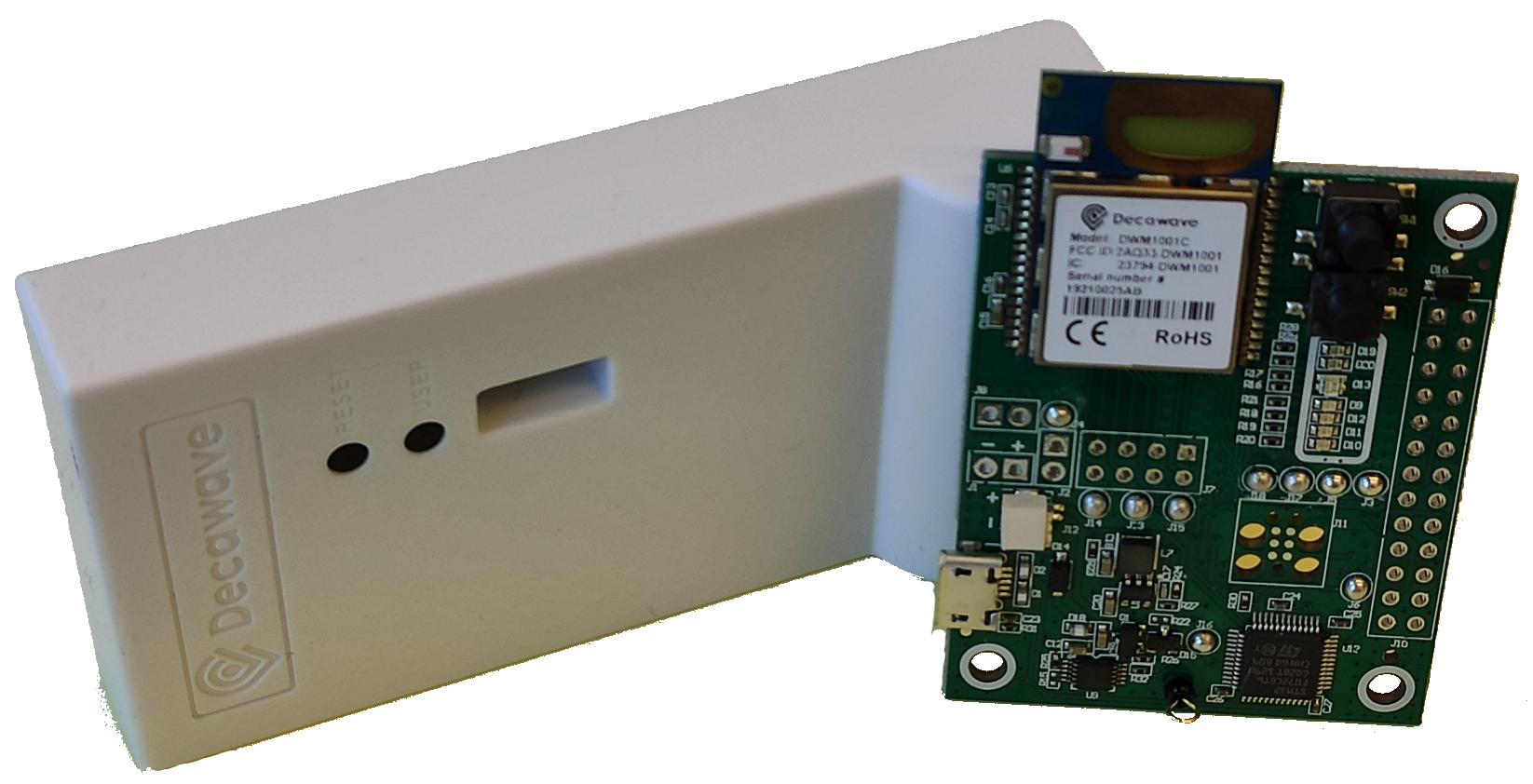} \\
        \footnotesize{(a) UWB Node.}
    \end{minipage}
    \begin{minipage}{0.24\textwidth}
        \centering
        \includegraphics[width=\textwidth, height=0.6\textwidth]{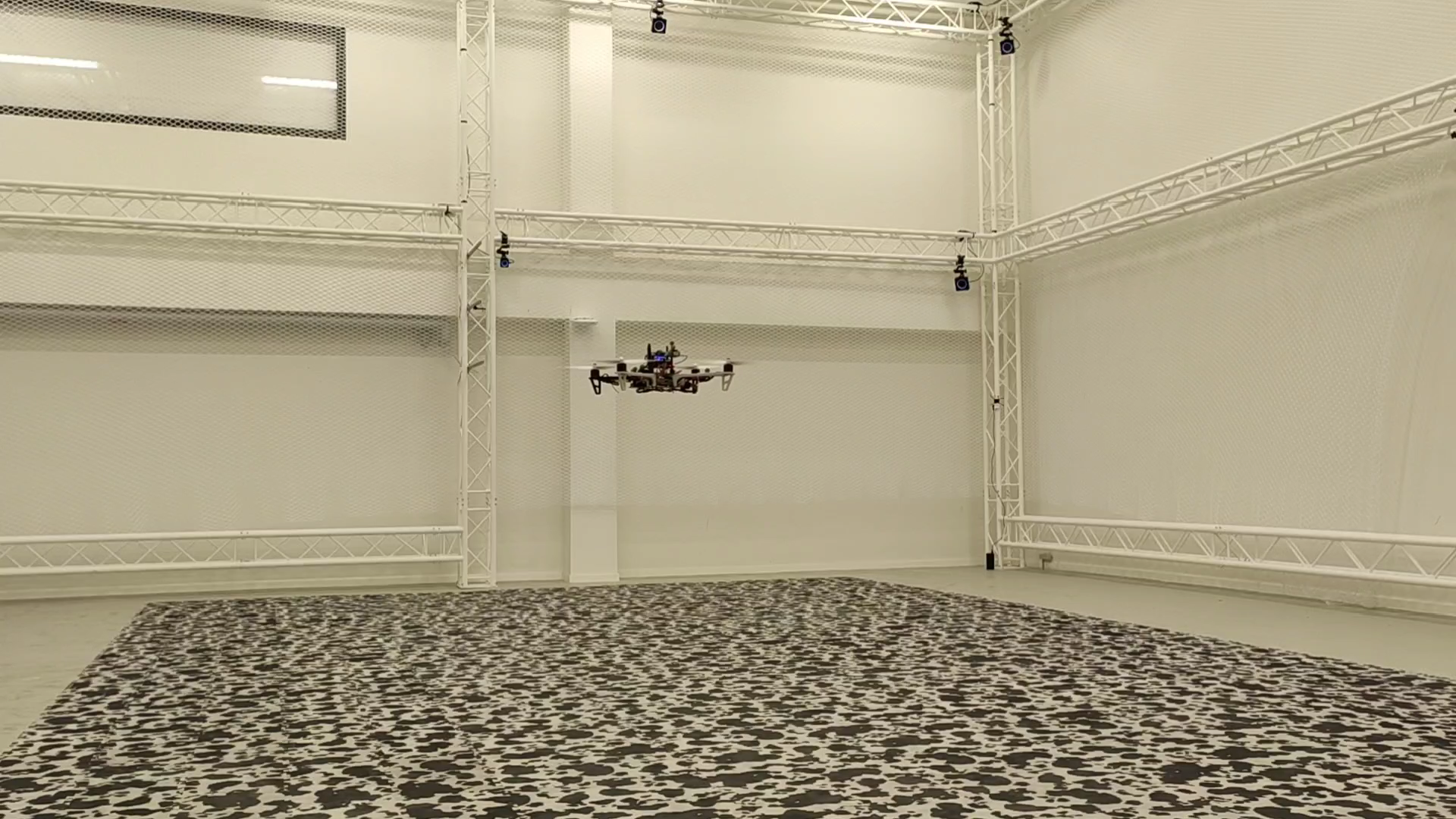} \\
        \footnotesize{(b) Quadrotor in the MOCAP system.}
    \end{minipage}
    \caption{(a) The DWM1001 DEV board with and without case. (b) The F450 quadrotor used in the experiments in the Optitrack MOCAP arena.}
    \label{fig:drone_and_dwm1001}
\end{figure}

\section{Related Works}

There are some previous works which have already characterized UWB localization systems~\cite{mimoune2019evaluation}, including in the field of aerial robots~\cite{macoir2019uwb}. However, to the best of our knowledge, previous analysis of the localization accuracy were done based on a fixed and well-calibrated anchor system. Moreover, existing datasets are small and very specific. Therefore, we believe there is a need for a more comprehensive understanding of the advantages and limitations of UWB-based flight for autonomous UAVs, in particular with fast ad-hoc deployments and movable anchor systems where the relative position of the anchors can also change over time. Through the rest of this section, we explore recent works in UWB-based or UWB-aided localization for mobile robots and, in particular, UAVs. This covers use cases in both industry and academia. Then, we analyze existing datasets for UWB-based localization of mobile robots and compare then with ours.

\subsection{UWB in Mobile and Aerial Robots}
UWB ranging has been used in multiple mobile robots to air localization or navigation, often when fused with data from other sensors. For instance, UWB has been utilized to bypass the complexity of visual loop closure detection in~\cite{wang2017ultra}. In lidar-based SLAM, UWB ranging has been utilized to avoid laser range limitations of inexpensive 2D scanners in tunnel-like environments~\cite{song2019uwb}, where the UWB measurements are not utilized for positioning but as part of range-only SLAM. This simplifies the installation, as the position of the UWB anchors can be unknown. 

In multi-robot systems, having multiple UWB nodes in each robot enables robust relative localization, both in position and orientation~\cite{nguyen2018robust}. In the case of multi-robot coordination with an anchor-based positioning system, UWB distance measurements can be utilized in formation control algorithms~\cite{qiang2018formation}, also fusing them with odometry estimation~\cite{guo2019ultra}. Finally, the positioning system can also be decentralized with individual node-to-node distance measurements in a cooperative multi-agent system~\cite{liu2017cooperative}.

Multiple works have utilized UWB-based localization systems to enable indoor UAV flights. In one of the earliest implementations of such system, Tiemann \textit{et al.} study the robustness of a predefined UAV flight that relies on UWB ranging~\cite{tiemann2017scalable}. This has potential applications in the logistics sector. For instance, in~\cite{tiemann2017scalable} the authors rely on UWB-based UAV navigation for fast and flexible automated stocktaking in a warehouse by scanning the good's QR-codes. A similar study for UWB-based autonomous flight in warehouses was carried out in~\cite{tiemann2018enhanced}. When mapping with UAVs, accurate UWB-based positioning allows to shift the focus from odometry and position estimation to sensing, enabling high-fidelity three-dimensional reconstruction with RGB-D cameras~\cite{perez2017multi}.

A more complex experiment was carried out in~\cite{nguyen2018integrated}, where the authors show how UWB-based localization can be combined with vision position estimation for docking UAVs on a ground vehicle. The work is extended in~\cite{nguyen2019integrated} with a focus on GNSS-denied environments. In their setting, four UWB anchor nodes are placed in the ground robot while one UWB tad is placed on the drone. However, the ground robot is relatively large, 2~m long and 1.5~m wide. In our dataset, we experiment with different anchor configurations, including a setting that can be installed on a small ground robot of about 0.6~m by 0.6~m, and a case where all four anchors are near to the ground.

\subsection{UWB Localization Datasets}

Raza \textit{et al.} introduced a dataset for indoor localization with narrow-band and ultra-wideband systems~\cite{raza2019dataset}. The dataset includes data from both a walking subject and a remotely operated radio control car. This dataset only one specific anchor position, and the UWB tag in the remotely operated car is at a constant height. In consequence, the analysis of the data can only be partly extrapolated to other use cases, such as aerial robots.

Barral \textit{et al.} presented a dataset acquired using ROS and Pozyx UWB devices~\cite{barral2019dataset}. This dataset only contains range information between two tags. It enables the characterization of inter-device distance estimation in both line of sight (LOS) and non-line of sight (NLOS) conditions. A similar study was carried out by Bregar \textit{et al.} in~\cite{bregar2016nlos} and~\cite{bregar2018dataset} with Decawave's DWM1000. In both cases, the distance between a single anchor and a tag was estimated in multiple locations, with both LOS and NLOS ranging.

Regarding the utilization of UWB localization for UAVs, Li \textit{et al.} published a dataset recorded over an indoor flight of a UAV with UWB-aided navigation~\cite{li2018dronedataset}. In their paper, the authors also introduce an Extended Kalman Filter (EKF) that enables very accurate 3D localization by fusing UWB and IMU data. Their dataset, however, contains data from a single flight with a single anchor setting. In contrast, our objective is to analyze how different anchor configurations affect the accuracy of the localization. In particular, most of our subsets of data have been recorded with anchors situated in a two-dimensional plane, so that it can mimic a more realistic and quick deployment in, for example, post-disaster scenarios, where the anchors might be mounted on ground robots. As most drones have some type of accurate onboard altitude estimation (sonar, lidar, or infrared), it is enough if the UWB system provides position information in two dimensions only. Moreover, we provide subsets of data were the quadrotor is equipped with one, two or four UWB tags, therefore enabling orientation estimation as well. Finally, commercially available UWB-based localization systems have significantly improved over the past two years since the previous dataset was published~\cite{li2018dronedataset}. In this paper, we report even smaller localization errors \textit{out of the box}, without the EKF to fuse with IMU data. The device we have utilized, the DWM1001 from Decawave, is illustrated in Fig.~\ref{fig:drone_and_dwm1001}\,(a).


\begin{table*}
    \centering
    \caption{Comparison of existing UWB-based localization and positioning datasets and ours.}
    \label{tab:datasets_comparison}
    \small
    \begin{tabular}{@{}lccccccccc@{}}
        \toprule
        & Distance   & Height     & Mobile & Mobile  & Anchor & Subsets & Convex   & Dims. & UWB  \\
        & Est. & Est. & tags   & anchors & settings & & Envelope &       & Node \\
        \midrule
        Bregar~\cite{bregar2016nlos} (2016)     & ToF   & -        & -  & -  & -  & - & - & 1D & DWM1000 \\
        Bregar~\cite{bregar2018dataset} (2018)  & ToF   & -        & -  & -  & -  & - & - & 1D & DWM1000 \\
        Raza~\cite{raza2019dataset} (2019)     & TDoA  & -        & \yess & - & 1 & 1 & IN & 2D & DWM1001 \\
        Barral~\cite{barral2019dataset} (2019)  & ToF   & -        & -  & -  & -  & - & - &1D & Pozyx \\
        Li~\cite{li2018dronedataset} (2018)     & TDoA  & UWB       & \yess & - & 1 & 1 & IN & 3D & TimeDomain \\
        \textbf{Ours}                    & ToF   & UWB+Lidar & \yess & \yess & 4 & 12 & IN/OUT & 3D & DWM1001 \\
        \bottomrule
    \end{tabular}
\end{table*}

Another key difference of our dataset is that we rely on ToF measurements only for the UWB position estimation. While this can reduce the accuracy when compared to TDoA, it does not require the anchors to be connected and synchronized. We believe this is an essential enabler of ad-hoc mobile deployments. A more detailed comparison of our dataset with existing ones is shown in Table~\ref{tab:datasets_comparison}.

\begin{figure}
    \centering
    \includegraphics[width=0.42\textwidth]{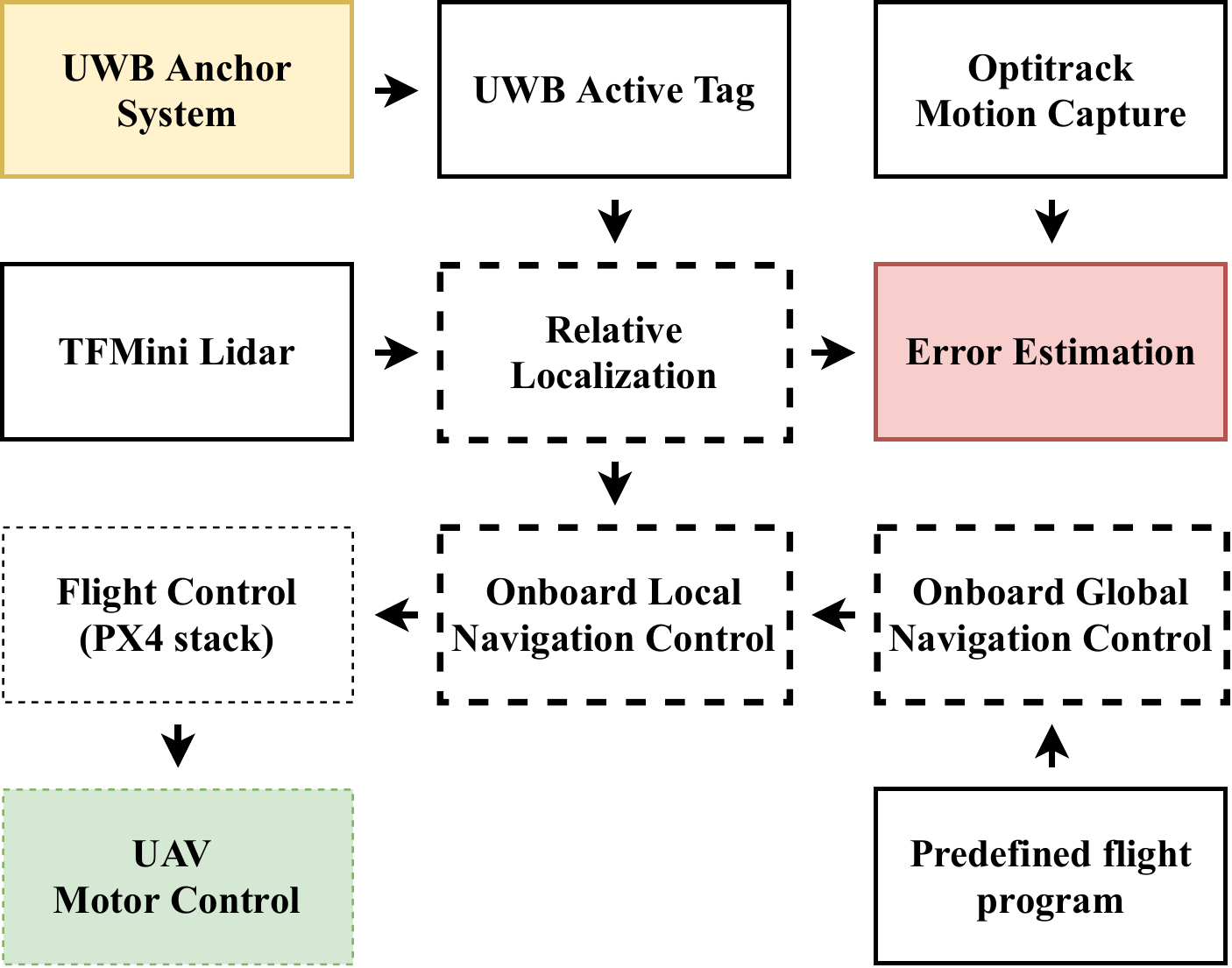}
    \caption{Controller blocks in the UWB-aided autonomous flight. The boxes with continuous border represent data acquisition ROS nodes, while dotted lines represent the custom ROS nodes where the actual control happens. The PX4 stack has not been modified.}
    \label{fig:rosnodes}
\end{figure}

\begin{table}
    \centering
    \caption{Latency and Accuracy of the Autopositioning method from Decawave's DRTLS localization system compared to a custom self-calibration method for anchors.}
    \begin{tabular}{@{}lccc@{}}
        \toprule
        & \textbf{Latency} & \textbf{Distance} & \textbf{Max. Error} \\ 
        \midrule
        \textbf{RTLS Autopositioning}       & $40\,s\pm5\,s$    & 10\,m     & 1.2\,m \\
                                            &                   & 4\,m     & 0.75\,m \\
        \textbf{Custom Calibration (x50)}   & $2.5\,s\pm0.1\,s$   & 10\,m     & 0.4\,m  \\
                                            &                   & 4\,m     & 0.25\,m \\
        \textbf{Custom Calibration (x5)}    & $0.9\,s\pm0.05\,s$  & 10\,m     & 0.5\,m  \\
                                            &                   & 4\,m     & 0.3\,\,m \\
        
        \bottomrule
    \end{tabular}
    \label{tab:autopositioning}
\end{table}

\begin{table*}
    \centering
    \caption{Description of the Data Subsets} 
    $ $\\[-16pt]
    \footnotesize{For each subset, we include the description of the anchor locations, the height at which the anchors are installed, the number of anchors and number of tags, the number of flights recorded separately, the total number of individual distance measurements or position estimations at 10~Hz (\#meas). In addition, we describe the platform where the data was recorded, whether at the companion computer connected to active tags, or at the base station via a passive tag, and whether the UAV flight happened inside or outside the convex envelope defined by the anchor positions.} \\[+1em]
    \label{tab:our_dataset}
    \small
    \begin{tabular}{@{}llccccccc@{}}
        \toprule
         & Anchor positions & Anchor & \#Anchors & \#Tags & \#Meas & \#Flights & Recording & UAV Position \\
         & & Height & & & & & (UAV/GS) & (w.r.t. Convex Env.) \\
        \midrule
        \#1 & Room corners  & 1.8~m       & 4 & 1 & 1000        & 1                 & UAV (Autonomous)   & Always in \\
        \#2 & Room corners  & 1.8~m       & 4 & 2 & 1000        & 1                 & UAV/GS   & Always in \\
        \#3 & Room corners  & 1.8~m       & \numrange{3}{4}     & 1                 & 9000    & 3     & GS    & Always in \\
        \#4 & Room corners  & 1.8~m       & \numrange{3}{4}     & 2                 & 7000    & 9     & GS    & Always in \\
        \#5 & Room corners  & \numrange{0}{2.4}~m  & 6          & 2                 & 1000     & 1     & UAV   & Always in \\
        \#6 & Single corner & 0 -- 0.5~m  & 4                   & 1                 & 1700 & 1     & GS    & Always out \\
        \#7 & Room center   & 0.1~m       & 4                   & \numrange{1}{4}   & 2500    & 3     & GS    & In \& Out \\
        \bottomrule
    \end{tabular}
\end{table*}

\section{Methodology}

This section describes the data that is included in the dataset as well as the steps followed when recording the different subsets. All the data, ROS nodes and firmware for the UWB devices is made publicly available in Github\footnote{\url{https://github.com/TIERS/UWB\_DRONE\_DATASET}}.

\subsection{UWB Ranging with ToF measurements}

UWB-based ranging between two unsynchronized devices is typically estimated through single-sided or double-sided two-way ranging (SS-TWR and DS-TWR, respectively), where the distance is calculated from the time of flight of an UWB signal query and its reply. In SS-TWR, the ToF is gien by \eqref{eq:ss-twr}:

\begin{equation}
    ToF=0.5*{\left ( T_{round}- T_{reply}\right )}
    \label{eq:ss-twr}
\end{equation}

where $T_{round}$ is the total time since the query signal is sent by an initiator node and until the reply signal is received, and $T_{reply}$ is the time it takes for the replying node to process the query and send the reply. This time is embedded on the reply for the initiator. In DS-TWR, the ToF is given by \eqref{eq:ds-twr}:

\begin{equation}
    ToF=\frac{T_{round1}*T_{round2}-T_{reply1}*T_{reply2}}{T_{round1}+T_{round2}+T_{reply1}+T_{reply2}}
    \label{eq:ds-twr}
\end{equation}

where now the initiator sends a second reply and it is the replying node that can calculate more accurately the ToF.

\subsection{Data Acquisition}

The dataset has been recorded using two different methods, with recordings on the UAV or at a ground station. The first case includes data which has been acquired using an onboard computer on a quadrotor equipped with an active UWB tag and flying autonomously. The second case refers to data from a passive tag connected to a ground station, while the quadrotor is being flown manually. In the second case, a delay exists between the ground truth data given by the motion capture system and the UWB data due to the passive nature of the tag being used for recording the positions.

The autonomous flight tests have been done with an F450 quadrotor equipped with a Pixhawk 2.4 running PX4, an Intel Up Board as a companion computer running Ubuntu 16.04 with ROS Kinetic and MAVROS, and a TF Mini Lidar for height estimation. The quadrotor flying in the motion capture arena (Optitrack system) is shown in Figure~\ref{fig:drone_and_dwm1001}~(b).

\subsection{Data Subsets}

The dataset presented in this paper contains 7 subsets listed in Table~\ref{tab:our_dataset}. In all cases, the number of anchors in use was either 3, 4 or 6. Four anchors is the minimum required for robust localization, as the position of a tag can still be calculated if the connection with one of the anchors is intermittent. However, at some points we disconnected one of the anchors to emulate the situations in which not all four anchors are reachable and study the impact on accuracy. Besides, while a larger amount of anchors can lead to an increase in accuracy, having ad-hoc and movable anchor networks with very large numbers might be impractical. In general, based on our experiments we believe that four anchors give enough accuracy to enable autonomous flight of an UAV, and therefore we find it the most suitable solution. The only scenario with 6 anchors was set in order to analyze if the vertical accuracy would change significantly.

The first four subsets were recorded in a more traditional setting with all anchors located in the corners of the motion capture arena. In the case of six anchors, the two extra ones were located on the walls at different heights from the original four. Then, the subset \#6 was recorded with four anchors in a single corner and near the ground, emulating a setting that could be installed onboard a single ground robot. Finally, subset \#7 was 
recorded with 4 anchors at a height of 10~cm and a separation of about 1.8~m. This configuration can be achieved with 4 small ground robots porting a UWB anchor each. While a bigger separation can result in better accuracy, our aim in this case was to test the localization estimation robustness when flying both inside and outside the convex envelope defined by the anchor positions.

Whenever more than one tag has been utilized, they were always part of a single rigid body. In subsets \#3 and \#4, two tags separated 30\,cm are mounted on top of the UAV, one in the front and one in the back. In the experiments with 4 tags, these were forming a rectangle of 20\,cm by 30\,cm. The position of the anchors is calculated through triangulation, and the measurements are taken at the maximum frequency allowed by the UWB module. This results in a higher frequency and larger number of measurements when compared to Decawave's solution, which is built to support a larger number of UWB nodes siultaneously.

\subsection{Dynamic anchor reconfiguration}

The public dataset that we have made available contains only data that has been acquired using Decawave's RTLS system with their proprietary firmware flashed onto the DWM1001 nodes. The product's UART API has been utilized as an interface to read distance and position information of the different nodes. However, we have found the RTLS calibration of anchor positions to be slow, taking around 40~s, and inaccurate, with errors exceeding 1~m in some cases. Therefore, we also make available an initial implementation of a custom anchor re-calibration system. Our system does between 5 and 50 measurements for each of the distances, with a total latency varying from just under 1~s to 2.5~s.

\subsection{Energy Efficiency}

We have also analyzed the power consumption of the UWB nodes in different modes. The power consumption has been monitored with Monsoon's High Voltage (HV) power monitor. While for UAVs the impact on total energy expenditure might be insignificant during flight, this can be a key aspect to take into consideration in mobile settings. For instance, if ground robots move only from time to time.

\begin{figure*}
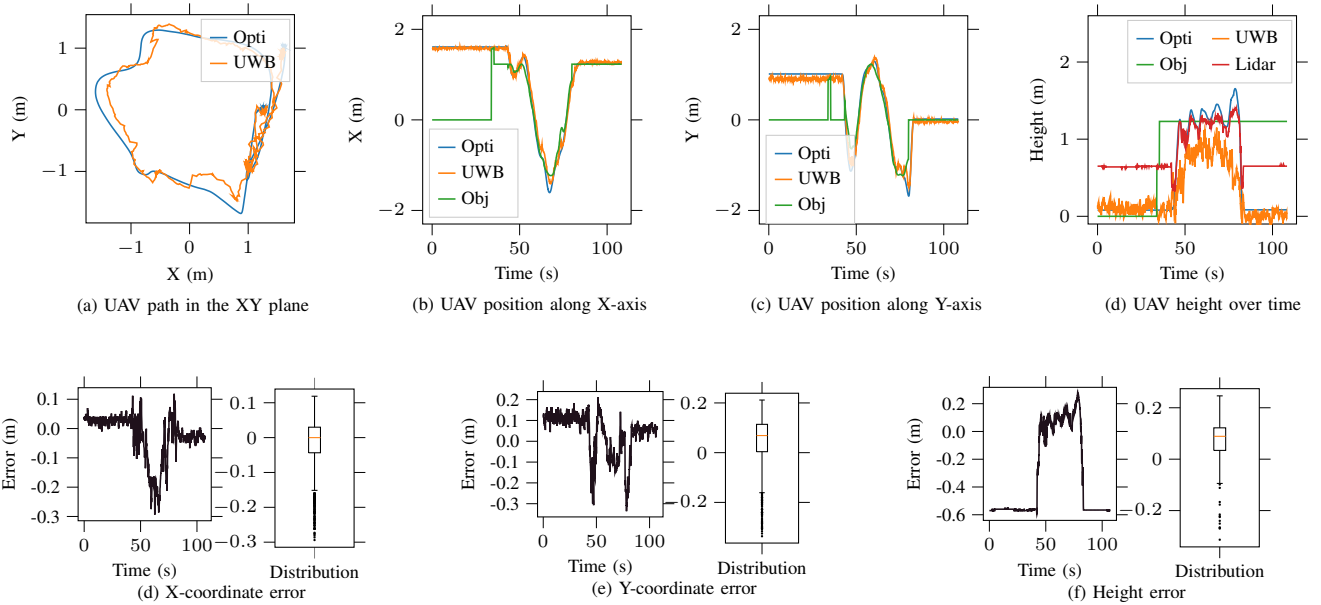

    \centering

    \begin{minipage}{0.24\textwidth}
        \centering
        \setlength{\figureheight}{\textwidth}
        \setlength{\figurewidth}{\textwidth}
        \scriptsize{\input{fig/tex/xy_paths.tex}} \\
        \hspace{0.9cm} (a) UAV path in the XY plane
    \end{minipage}
    \begin{minipage}{0.24\textwidth}
        \centering
        \setlength{\figureheight}{\textwidth}
        \setlength{\figurewidth}{\textwidth}
        \scriptsize{\input{fig/tex/x_trajectory.tex}} \\
        \hspace{0.9cm} (b) UAV position along X-axis
    \end{minipage}
    \begin{minipage}{0.24\textwidth}
        \centering
        \setlength{\figureheight}{\textwidth}
        \setlength{\figurewidth}{\textwidth}
        \scriptsize{\input{fig/tex/y_trajectory.tex}} \\
        \hspace{0.9cm} (c) UAV position along Y-axis
    \end{minipage}
    \begin{minipage}{0.24\textwidth}
        \centering
        \setlength{\figureheight}{\textwidth}
        \setlength{\figurewidth}{\textwidth}
        \scriptsize{\input{fig/tex/height.tex}} \\
        \hspace{0.9cm} (d) UAV height over time
    \end{minipage}
    
    $ $ \\[+23pt]
    \begin{minipage}{0.33\textwidth}
        \begin{minipage}{0.56\textwidth}
            \centering
            \setlength{\figureheight}{\textwidth}
            \setlength{\figurewidth}{\textwidth}
            \scriptsize{\input{fig/tex/x_error.tex}} \\
        \end{minipage}\hspace{-0.72cm}
        \begin{minipage}{0.44\textwidth}
            \centering
            \setlength{\figureheight}{1.4\textwidth}
            \setlength{\figurewidth}{\textwidth}
            \scriptsize{
\begin{tikzpicture}

\definecolor{color0}{rgb}{1,0.498039215686275,0.0549019607843137}

\begin{axis}[
height=\figureheight,
tick align=outside,
width=\figurewidth,
x grid style={lightgray!92.02614379084967!black},
xmin=0.5, xmax=1.5,
xtick={1},
xticklabels={Distribution},
y grid style={lightgray!92.02614379084967!black},
ymin=-0.3145841922765, ymax=0.1393631639485,
ytick style={color=black},
ytick={-0.3,-0.2,-0.1,0,0.1},
]
\addplot [black]
table {%
0.925 -0.0435044693955
1.075 -0.0435044693955
1.075 0.02990476608
0.925 0.02990476608
0.925 -0.0435044693955
};
\addplot [black]
table {%
1 -0.0435044693955
1 -0.15134084702
};
\addplot [black]
table {%
1 0.02990476608
1 0.118729193211
};
\addplot [black]
table {%
0.9625 -0.15134084702
1.0375 -0.15134084702
};
\addplot [black]
table {%
0.9625 0.118729193211
1.0375 0.118729193211
};
\addplot [black, mark=*, mark size=0.23, mark options={solid,fill opacity=0}, only marks]
table {%
1 -0.15965015173
1 -0.166317597628
1 -0.16271420002
1 -0.162456445694
1 -0.169286949635
1 -0.162556141019
1 -0.181938742399
1 -0.181844568253
1 -0.181769020557
1 -0.187282958031
1 -0.1974044260383
1 -0.1714503069222
1 -0.1884974397719
1 -0.18715131871402
1 -0.1843051159382
1 -0.1848713091016
1 -0.194952856898
1 -0.175120823979
1 -0.174236009121
1 -0.163954080343
1 -0.182774401307
1 -0.21170391798
1 -0.235593539476
1 -0.232390426397
1 -0.275033774376
1 -0.261308455467
1 -0.236632442474
1 -0.251200940609
1 -0.245377210379
1 -0.228582818508
1 -0.212129871845
1 -0.165091199875
1 -0.176845395565
1 -0.16937517643
1 -0.176493442059
1 -0.17927970171
1 -0.197038300037
1 -0.180557408333
1 -0.18295260191
1 -0.173279573917
1 -0.172363312244
1 -0.169882249832
1 -0.185738694668
1 -0.179870538712
1 -0.184828503132
1 -0.174309277534
1 -0.182789289951
1 -0.162789289951
1 -0.170359973907
1 -0.189499680996
1 -0.187445218563
1 -0.199813966751
1 -0.184638962746
1 -0.166552391052
1 -0.208771591187
1 -0.261029832363
1 -0.293950221539
1 -0.245805058479
1 -0.217004363537
1 -0.211866195202
1 -0.206304535866
1 -0.189604845047
1 -0.174810297489
1 -0.170779047012
1 -0.169307875633
1 -0.171090819836
1 -0.181318440437
1 -0.186405389309
1 -0.19410492897
1 -0.203458497524
1 -0.215226829052
1 -0.23196957588
1 -0.20655710697
1 -0.19572083473
1 -0.21988311768
1 -0.22538413048
1 -0.21624098778
1 -0.21844954967
1 -0.21623210907
1 -0.21432556629
1 -0.21286176682
1 -0.21177788734
1 -0.25151726723
1 -0.26153559685
1 -0.2313618803
1 -0.22499726772
1 -0.23806652069
1 -0.2443209362
1 -0.25852459431
1 -0.26134375572
1 -0.27295890331
1 -0.28642012596
1 -0.28115880489
1 -0.25115880489
1 -0.27714689732
1 -0.26207071781
1 -0.25430182934
1 -0.24213268757
1 -0.22713423252
1 -0.24228276253
1 -0.23459160805
1 -0.24601437092
1 -0.20562670231
1 -0.20223042965
1 -0.19922754765
1 -0.1827271843
1 -0.17456718922
1 -0.17224714756
1 -0.16027472019
1 -0.15977202415
1 -0.160127239227
1 -0.188506836891
1 -0.17755294323
1 -0.188419246674
1 -0.183808300495
1 -0.177425844669
1 -0.176393561363
1 -0.184498143196
1 -0.192625911236
1 -0.212746798992
1 -0.182232480049
1 -0.192623734474
};
\addplot [color0]
table {%
0.925 -9.41014300001042e-05
1.075 -9.41014300001042e-05
};
\end{axis}

\end{tikzpicture}}
        \end{minipage}
        
        \centering
        \hspace{0.23cm} \scriptsize{(d) X-coordinate error}
    \end{minipage}
    \begin{minipage}{0.32\textwidth}
        \begin{minipage}{0.56\textwidth}
            \centering
            \setlength{\figureheight}{\textwidth}
            \setlength{\figurewidth}{\textwidth}
            \scriptsize{\input{fig/tex/y_error.tex}} \\
        \end{minipage}\hspace{-0.72cm}
        \begin{minipage}{0.44\textwidth}
            \centering
            \setlength{\figureheight}{1.4\textwidth}
            \setlength{\figurewidth}{\textwidth}
            \scriptsize{
\begin{tikzpicture}

\definecolor{color0}{rgb}{1,0.498039215686275,0.0549019607843137}

\begin{axis}[
height=\figureheight,
tick align=outside,
width=\figurewidth,
x grid style={lightgray!92.02614379084967!black},
xmin=0.5, xmax=1.5,
xtick style={color=black},
xtick={1},
xticklabels={Distribution},
y grid style={lightgray!92.02614379084967!black},
ymin=-0.3635190824825, ymax=0.2390229169725,
ytick style={color=black},
]
\addplot [black]
table {%
0.925 0.00392535745975001
1.075 0.00392535745975001
1.075 0.11388812065
0.925 0.11388812065
0.925 0.00392535745975001
};
\addplot [black]
table {%
1 0.00392535745975001
1 -0.16095939159
};
\addplot [black]
table {%
1 0.11388812065
1 0.21163464427
};
\addplot [black]
table {%
0.9625 -0.16095939159
1.0375 -0.16095939159
};
\addplot [black]
table {%
0.9625 0.21163464427
1.0375 0.21163464427
};
\addplot [black, mark=*, mark size=0.23, mark options={solid,fill opacity=0}, only marks]
table {%
1 -0.170456328392
1 -0.162841489315
1 -0.2037337327
1 -0.249427547455
1 -0.228966834545
1 -0.194187994003
1 -0.187495963573
1 -0.20815598488
1 -0.25503983498
1 -0.2992868948
1 -0.16267565727
1 -0.26902498245
1 -0.22662935257
1 -0.28174201012
1 -0.28561989784
1 -0.30720539093
1 -0.23713435173
1 -0.178700654507
1 -0.181291329861
1 -0.19382733345
1 -0.19242446899
1 -0.18837575436
1 -0.22654063702
1 -0.2431864357
1 -0.22971111774
1 -0.24082834721
1 -0.24082834721
1 -0.24082834721
1 -0.28330334187
1 -0.29815658092
1 -0.30555589199
1 -0.30529433727
1 -0.33613080978
1 -0.32503966808
1 -0.29449450493
1 -0.30132387638
1 -0.3158857584
1 -0.28710894585
1 -0.27883328915
1 -0.27670192719
1 -0.2428832674
1 -0.21409417629
1 -0.19921038151
1 -0.17015057564
1 -0.17816202641
1 -0.24034866333
1 -0.26331088543
1 -0.27362141609
1 -0.25099570274
1 -0.20184399605
1 -0.2175457859
1 -0.22942168236
1 -0.2481303072
1 -0.18090269566
1 -0.20840839863
1 -0.19395288944
1 -0.20050467968
1 -0.1830142498
};
\addplot [color0]
table {%
0.925 0.06873652637005
1.075 0.06873652637005
};
\end{axis}

\end{tikzpicture}}
        \end{minipage}
        
        \centering
        \hspace{0.23cm} \scriptsize{(e) Y-coordinate error}
    \end{minipage}
    \begin{minipage}{0.33\textwidth}
        \begin{minipage}{0.56\textwidth}
            \centering
            \setlength{\figureheight}{\textwidth}
            \setlength{\figurewidth}{\textwidth}
            \scriptsize{\input{fig/tex/z_error_lidar.tex}} \\
        \end{minipage}\hspace{-0.72cm}
        \begin{minipage}{0.44\textwidth}
            \centering
            \setlength{\figureheight}{1.4\textwidth}
            \setlength{\figurewidth}{\textwidth}
            \scriptsize{
\begin{tikzpicture}

\definecolor{color0}{rgb}{1,0.498039215686275,0.0549019607843137}

\begin{axis}[
height=\figureheight,
tick align=outside,
width=\figurewidth,
x grid style={lightgray!92.02614379084967!black},
xmin=0.5, xmax=1.5,
xtick style={color=black},
xtick={1},
xticklabels={Distribution},
y grid style={lightgray!92.02614379084967!black},
ymin=-0.34127654619506, ymax=0.27425946033786,
ytick style={color=black},
]
\addplot [black]
table {%
0.925 0.0341100394775
1.075 0.0341100394775
1.075 0.1219157278525
0.925 0.1219157278525
0.925 0.0341100394775
};
\addplot [black]
table {%
1 0.0341100394775
1 -0.094377875329
};
\addplot [black]
table {%
1 0.1219157278525
1 0.24628055095
};
\addplot [black]
table {%
0.9625 -0.094377875329
1.0375 -0.094377875329
};
\addplot [black]
table {%
0.9625 0.24628055095
1.0375 0.24628055095
};
\addplot [black, mark=*, mark size=0.23, mark options={solid,fill opacity=0}, only marks]
table {%
1 -0.3132976368072
1 -0.269783310592
1 -0.267545737326
1 -0.225128360093
1 -0.26535371691
1 -0.215293034911
1 -0.250682786107
1 -0.239843606949
1 -0.221727654338
1 -0.166557058692
1 -0.176819071174
1 -0.111699938779
1 -0.0986179113369999
};
\addplot [color0]
table {%
0.925 0.0889423787615
1.075 0.0889423787615
};
\end{axis}

\end{tikzpicture}}
        \end{minipage}
        
        \centering
        \hspace{0.23cm} \scriptsize{(f) Height error}
    \end{minipage}

    \caption{Autonomous flight based on UWB for localization in the XY plane and a 1D lidar for height estimation. For the height error, only the lidar data was taken into account, and only while on flight for the boxplot. The Optitrack (Opti) system gives the ground truth reference, while the position is estimated based on the UWB system and 1D lidar. The position control input (Obj) utilizes the position estimation as well.}
    \label{fig:circle}
    
\end{figure*}



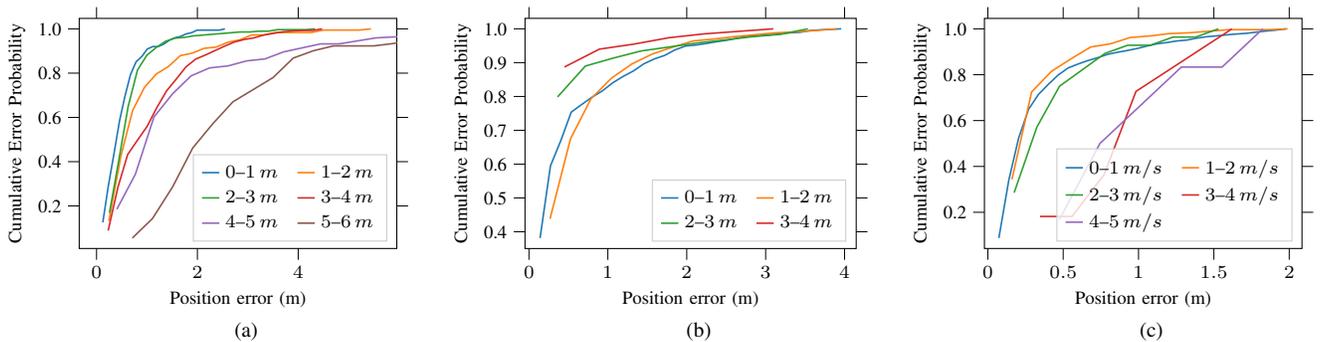
\begin{figure*}
    \centering
    \begin{minipage}{0.32\textwidth}
        \centering
        \setlength{\figureheight}{0.8\textwidth}
        \setlength{\figurewidth}{\textwidth}
        \scriptsize{
\begin{tikzpicture}

\definecolor{color1}{rgb}{1,0.498039215686275,0.0549019607843137}
\definecolor{color5}{rgb}{0.549019607843137,0.337254901960784,0.294117647058824}
\definecolor{color2}{rgb}{0.172549019607843,0.627450980392157,0.172549019607843}
\definecolor{color3}{rgb}{0.83921568627451,0.152941176470588,0.156862745098039}
\definecolor{color0}{rgb}{0.12156862745098,0.466666666666667,0.705882352941177}
\definecolor{color4}{rgb}{0.580392156862745,0.403921568627451,0.741176470588235}

\begin{axis}[
    height=\figureheight,
    legend cell align={left},
    legend style={fill opacity=0.8, draw opacity=1, text opacity=1, draw=white!80.0!black}, 
    legend pos=south east,
    tick align=outside,
    width=\figurewidth,
    x grid style={white!69.01960784313725!black},
    xlabel={Position error (m)},
    xmin=-0.343820904594427, xmax=5.95557510971825,
    xtick style={color=black},
    y grid style={white!69.01960784313725!black},
    ylabel={Cumulative Error Probability},
    ymin=0.00769230769230767, ymax=1.04725274725275,
    ytick style={color=black},
    ytick={0,0.2,0.4,0.6,0.8,1,1.2},
    yticklabels={0.0,0.2,0.4,0.6,0.8,1.0,1.2},
    legend columns=2, 
    /tikz/column 2/.style={
        column sep=5pt,
    },
    legend image code/.code={
        \draw[mark repeat=2,mark phase=2]
        plot coordinates {
            (0cm,0cm)
            (0.1cm,0cm) 
            (0.23cm,0cm)  
        };%
    },
]
\addplot [semithick, color0]
table {%
0.124333459692513 0.125714285714286
0.234399502196604 0.291428571428571
0.344465544700695 0.434285714285714
0.454531587204787 0.582857142857143
0.564597629708878 0.697142857142857
0.674663672212969 0.794285714285714
0.784729714717061 0.851428571428572
0.894795757221152 0.874285714285714
1.00486179972524 0.908571428571429
1.11492784222933 0.92
1.22499388473343 0.92
1.33505992723752 0.931428571428572
1.44512596974161 0.948571428571429
1.5551920122457 0.96
1.66525805474979 0.965714285714286
1.77532409725388 0.977142857142858
1.88539013975797 0.982857142857143
1.99545618226207 0.994285714285715
2.10552222476616 0.994285714285715
2.21558826727025 0.994285714285715
2.32565430977434 0.994285714285715
2.43572035227843 0.994285714285715
2.54578639478252 1
};
\addlegendentry{$\numrange{0}{1}\,m$}
\addplot [semithick, color1]
table {%
0.243735576456724 0.131868131868132
0.480028663675408 0.423076923076923
0.716321750894092 0.631868131868132
0.952614838112777 0.736263736263736
1.18890792533146 0.796703296703297
1.42520101255015 0.82967032967033
1.66149409976883 0.879120879120879
1.89778718698751 0.89010989010989
2.1340802742062 0.912087912087912
2.37037336142488 0.917582417582418
2.60666644864357 0.93956043956044
2.84295953586225 0.950549450549451
3.07925262308094 0.972527472527473
3.31554571029962 0.972527472527473
3.5518387975183 0.983516483516484
3.78813188473699 0.983516483516484
4.02442497195567 0.983516483516484
4.26071805917436 0.983516483516484
4.49701114639304 0.994505494505495
4.73330423361173 0.994505494505495
4.96959732083041 0.994505494505495
5.20589040804909 0.994505494505495
5.44218349526778 1
};
\addlegendentry{$\numrange{1}{2}\,m$}
\addplot [semithick, color2]
table {%
0.255981173225097 0.168421052631579
0.441344718505142 0.407017543859649
0.626708263785186 0.649122807017544
0.812071809065231 0.814035087719298
0.997435354345276 0.880701754385965
1.18279889962532 0.915789473684211
1.36816244490536 0.943859649122807
1.55352599018541 0.957894736842105
1.73888953546545 0.96140350877193
1.9242530807455 0.968421052631579
2.10961662602554 0.971929824561404
2.29498017130559 0.975438596491228
2.48034371658563 0.978947368421053
2.66570726186568 0.982456140350877
2.85107080714572 0.985964912280702
3.03643435242577 0.985964912280702
3.22179789770581 0.989473684210526
3.40716144298586 0.989473684210526
3.5925249882659 0.996491228070175
3.77788853354595 0.996491228070175
3.96325207882599 0.996491228070175
4.14861562410603 0.996491228070175
4.33397916938608 1
};
\addlegendentry{$\numrange{2}{3}\,m$}
\addplot [semithick, color3]
table {%
0.230713011863182 0.0885608856088561
0.423622063687807 0.284132841328413
0.616531115512432 0.431734317343174
0.809440167337057 0.496309963099631
1.00234921916168 0.559040590405904
1.19525827098631 0.643911439114391
1.38816732281093 0.719557195571956
1.58107637463556 0.773062730627306
1.77398542646018 0.828413284132842
1.96689447828481 0.863468634686347
2.15980353010943 0.881918819188192
2.35271258193406 0.902214022140222
2.54562163375868 0.922509225092251
2.73853068558331 0.940959409594096
2.93143973740793 0.952029520295203
3.12434878923256 0.959409594095941
3.31725784105718 0.972324723247233
3.51016689288181 0.979704797047971
3.70307594470643 0.987084870848709
3.89598499653106 0.990774907749078
4.08889404835568 0.992619926199262
4.2818031001803 0.992619926199262
4.47471215200493 1
};
\addlegendentry{$\numrange{3}{4}\,m$}
\addplot [semithick, color4]
table {%
0.404044272368085 0.18552036199095
0.77116334260014 0.343891402714932
1.1382824128322 0.601809954751131
1.50540148306425 0.705882352941176
1.87252055329631 0.787330316742081
2.23963962352836 0.823529411764706
2.60675869376042 0.832579185520362
2.97387776399247 0.855203619909502
3.34099683422453 0.864253393665158
3.70811590445658 0.895927601809955
4.07523497468864 0.914027149321267
4.4423540449207 0.932126696832579
4.80947311515275 0.932126696832579
5.17659218538481 0.945701357466063
5.54371125561686 0.959276018099548
5.91083032584892 0.963800904977376
6.27794939608097 0.972850678733032
6.64506846631303 0.981900452488688
7.01218753654508 0.986425339366516
7.37930660677714 0.990950226244344
7.74642567700919 0.995475113122172
8.11354474724125 0.995475113122172
8.48066381747331 1
};
\addlegendentry{$\numrange{4}{5}\,m$}
\addplot [semithick, color5]
table {%
0.710066775314246 0.054945054945055
1.10903741031957 0.142857142857143
1.50800804532489 0.285714285714286
1.90697868033021 0.461538461538462
2.30594931533553 0.571428571428571
2.70491995034085 0.67032967032967
3.10389058534617 0.725274725274725
3.50286122035149 0.78021978021978
3.90183185535681 0.868131868131868
4.30080249036214 0.901098901098901
4.69977312536746 0.923076923076923
5.09874376037278 0.923076923076923
5.4977143953781 0.923076923076923
5.89668503038342 0.934065934065934
6.29565566538874 0.956043956043956
6.69462630039406 0.967032967032967
7.09359693539938 0.978021978021978
7.4925675704047 0.978021978021978
7.89153820541003 0.978021978021978
8.29050884041535 0.989010989010989
8.68947947542067 0.989010989010989
9.08845011042599 0.989010989010989
9.48742074543131 1
};
\addlegendentry{$\numrange{5}{6}\,m$}
\end{axis}

\end{tikzpicture}} \\
        \hspace{1cm} \footnotesize{(a)}
    \end{minipage}
    \begin{minipage}{0.33\textwidth}
        \centering
        \setlength{\figureheight}{0.776\textwidth}
        \setlength{\figurewidth}{\textwidth}
        \scriptsize{
\begin{tikzpicture}

\definecolor{color3}{rgb}{0.83921568627451,0.152941176470588,0.156862745098039}
\definecolor{color1}{rgb}{1,0.498039215686275,0.0549019607843137}
\definecolor{color0}{rgb}{0.12156862745098,0.466666666666667,0.705882352941177}
\definecolor{color2}{rgb}{0.172549019607843,0.627450980392157,0.172549019607843}

\begin{axis}[
    height=\figureheight,
    legend cell align={left},
    legend style={fill opacity=0.8, draw opacity=1, text opacity=1, draw=white!80.0!black}, 
    legend pos=south east,
    tick align=outside,
    width=\figurewidth,
    x grid style={lightgray!92.02614379084967!black},
    xlabel={Position error (m)},
    xmin=-0.0498907200294144, xmax=4.147631825822,
    xtick style={color=black},
    y grid style={lightgray!92.02614379084967!black},
    ylabel={Cumulative Error Probability},
    ymin=0.350337078651685, ymax=1.03093632958802,
    ytick style={color=black},
    ytick={0.3,0.4,0.5,0.6,0.7,0.8,0.9,1,1.1},
    yticklabels={0.3,0.4,0.5,0.6,0.7,0.8,0.9,1.0,1.1},
    legend columns=2, 
    /tikz/column 2/.style={
        column sep=5pt,
    },
    legend image code/.code={
        \draw[mark repeat=2,mark phase=2]
        plot coordinates {
            (0cm,0cm)
            (0.1cm,0cm) 
            (0.23cm,0cm)  
        };%
    },
]
\addplot [semithick, color0]
table {%
0.140905759327468 0.3812734082397
0.272489538194284 0.594756554307116
0.404073317061099 0.669662921348315
0.535657095927915 0.753558052434457
0.66724087479473 0.776029962546816
0.798824653661546 0.797752808988764
0.930408432528361 0.816479400749064
1.06199221139518 0.840449438202247
1.19357599026199 0.859176029962547
1.32515976912881 0.875655430711611
1.45674354799562 0.896629213483146
1.58832732686244 0.910861423220974
1.71991110572925 0.921348314606742
1.85149488459607 0.937827715355805
1.98307866346289 0.949063670411985
2.1146624423297 0.952059925093633
2.24624622119652 0.956554307116105
2.37783000006333 0.961797752808989
2.50941377893015 0.966292134831461
2.64099755779696 0.972284644194757
2.77258133666378 0.978277153558053
2.90416511553059 0.979775280898877
3.03574889439741 0.982022471910112
3.16733267326423 0.984269662921348
3.29891645213104 0.988014981273408
3.43050023099786 0.989513108614232
3.56208400986467 0.994756554307116
3.69366778873149 0.997003745318352
3.8252515675983 0.998501872659176
3.95683534646512 1
};
\addlegendentry{$\numrange{0}{1}\,m$}
\addplot [semithick, color1]
table {%
0.266985861962144 0.438288920056101
0.526529234692821 0.674614305750351
0.786072607423497 0.79312762973352
1.04561598015417 0.85413744740533
1.30515935288485 0.896213183730715
1.56470272561553 0.925666199158485
1.8242460983462 0.946002805049088
2.08378947107688 0.964235624123422
2.34333284380756 0.96984572230014
2.60287621653823 0.976858345021038
2.86241958926891 0.982468443197756
3.12196296199959 0.988078541374474
3.38150633473026 0.990883590462833
3.64104970746094 0.995091164095372
3.90059308019162 1
};
\addlegendentry{$\numrange{1}{2}\,m$}
\addplot [semithick, color2]
table {%
0.36249318267955 0.798594847775176
0.715052287988805 0.889929742388759
1.06761139329806 0.913348946135831
1.42017049860731 0.934426229508197
1.77272960391657 0.946135831381733
2.12528870922583 0.957845433255269
2.47784781453508 0.967213114754098
2.83040691984434 0.976580796252927
3.18296602515359 0.983606557377049
3.53552513046285 1
};
\addlegendentry{$\numrange{2}{3}\,m$}
\addplot [semithick, color3]
table {%
0.450936296291633 0.887218045112782
0.891972571994114 0.93984962406015
1.3330088476966 0.954887218045113
1.77404512339908 0.973684210526316
2.21508139910156 0.984962406015038
2.65611767480404 0.992481203007519
3.09715395050652 1
};
\addlegendentry{$\numrange{3}{4}\,m$}
\end{axis}

\end{tikzpicture}} \\
        \hspace{1cm} \footnotesize{(b)}
    \end{minipage}
    \begin{minipage}{0.32\textwidth}
        \centering
        \setlength{\figureheight}{0.8\textwidth}
        \setlength{\figurewidth}{\textwidth}
        \scriptsize{
\begin{tikzpicture}

\definecolor{color1}{rgb}{1,0.498039215686275,0.0549019607843137}
\definecolor{color3}{rgb}{0.83921568627451,0.152941176470588,0.156862745098039}
\definecolor{color0}{rgb}{0.12156862745098,0.466666666666667,0.705882352941177}
\definecolor{color2}{rgb}{0.172549019607843,0.627450980392157,0.172549019607843}
\definecolor{color4}{rgb}{0.580392156862745,0.403921568627451,0.741176470588235}

\begin{axis}[
    height=\figureheight,
    legend cell align={left},
    legend style={fill opacity=0.8, draw opacity=1, text opacity=1, draw=white!80.0!black}, 
    legend pos=south east,
    tick align=outside,
    width=\figurewidth,
    x grid style={lightgray!92.02614379084967!black},
    xlabel={Position error (m)},
    xmin=-0.0226749965894595, xmax=2.08480568253259,
    xtick style={color=black},
    y grid style={lightgray!92.02614379084967!black},
    ylabel={Cumulative Error Probability},
    ymin=0.0420676998368678, ymax=1.04561582381729,
    ytick style={color=black},
    ytick={0,0.2,0.4,0.6,0.8,1,1.2},
    yticklabels={0.0,0.2,0.4,0.6,0.8,1.0,1.2},
    legend columns=2, 
    /tikz/column 2/.style={
        column sep=5pt,
    },
    legend image code/.code={
        \draw[mark repeat=2,mark phase=2]
        plot coordinates {
            (0cm,0cm)
            (0.1cm,0cm) 
            (0.23cm,0cm)  
        };%
    },
]
\addplot [semithick, color0]
table {%
0.0731195797342698 0.0876835236541599
0.138796670237072 0.33931484502447
0.204473760739875 0.524061990212072
0.270150851242677 0.648858075040783
0.33582794174548 0.714110929853181
0.401505032248282 0.75815660685155
0.467182122751084 0.798939641109298
0.532859213253887 0.82952691680261
0.598536303756689 0.846655791190865
0.664213394259492 0.861745513866232
0.729890484762294 0.874388254486134
0.795567575265097 0.889070146818923
0.861244665767899 0.898042414355628
0.926921756270701 0.9057911908646
0.992598846773504 0.914763458401305
1.05827593727631 0.92536704730832
1.12395302777911 0.93515497553018
1.18963011828191 0.940456769983687
1.25530720878471 0.947389885807504
1.32098429928752 0.951060358890702
1.38666138979032 0.959624796084829
1.45233848029312 0.965742251223491
1.51801557079592 0.969412724306689
1.58369266129873 0.973898858075041
1.64936975180153 0.977569331158238
1.71504684230433 0.980424143556281
1.78072393280713 0.985725938009788
1.84640102330994 0.990619902120718
1.91207811381274 0.994698205546493
1.97775520431554 1
};
\addlegendentry{$\numrange{0}{1}\,m/s$}
\addplot [semithick, color1]
table {%
0.160055242309858 0.343642611683849
0.290694946874073 0.725085910652921
0.421334651438287 0.814432989690722
0.551974356002501 0.869415807560137
0.682614060566715 0.920962199312715
0.813253765130929 0.934707903780069
0.943893469695143 0.962199312714777
1.07453317425936 0.969072164948453
1.20517287882357 0.979381443298969
1.33581258338779 0.982817869415808
1.466452287952 0.989690721649484
1.59709199251621 0.996563573883161
1.72773169708043 0.996563573883161
1.85837140164464 0.996563573883161
1.98901110620886 1
};
\addlegendentry{$\numrange{1}{2}\,m/s$}
\addplot [semithick, color2]
table {%
0.174683212903234 0.285714285714286
0.325284852928694 0.571428571428571
0.475886492954154 0.75
0.626488132979613 0.821428571428571
0.777089773005073 0.892857142857143
0.927691413030533 0.928571428571428
1.07829305305599 0.928571428571428
1.22889469308145 0.964285714285714
1.37949633310691 0.964285714285714
1.53009797313237 1
};
\addlegendentry{$\numrange{2}{3}\,m/s$}
\addplot [semithick, color3]
table {%
0.346129612644636 0.181818181818182
0.558459399049607 0.181818181818182
0.770789185454578 0.363636363636364
0.983118971859548 0.727272727272727
1.19544875826452 0.818181818181818
1.40777854466949 0.909090909090909
1.62010833107446 1
};
\addlegendentry{$\numrange{3}{4}\,m/s$}
\addplot [semithick, color4]
table {%
0.473396784673401 0.166666666666667
0.743664452958496 0.5
1.01393212124359 0.666666666666667
1.28419978952868 0.833333333333333
1.55446745781378 0.833333333333333
1.82473512609887 1
};
\addlegendentry{$\numrange{4}{5}\,m/s$}
\end{axis}

\end{tikzpicture}} \\
        \hspace{1cm} \footnotesize{(c)}
    \end{minipage}
    \caption{Cumulative probability distribution of the positioning accuracy for (a) different distances to the center of mass of the anchor system, (b) different heights, (c) and different speeds of the UAV. The anchors are positioned in the room corners forming a square of $64\,m^2$.}
    \label{fig:corners_cumulatives}
\end{figure*}

\begin{figure}
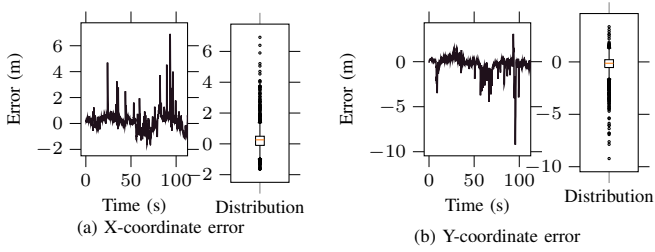

    \centering
    \begin{minipage}{0.24\textwidth}
        \begin{minipage}{0.56\textwidth}
            \centering
            \setlength{\figureheight}{1.375\textwidth}
            \setlength{\figurewidth}{1.23\textwidth}
            \scriptsize{\input{fig/one_corner/x_error.tex}} \\
        \end{minipage}\hspace{-0.23cm}
        \begin{minipage}{0.44\textwidth}
            \centering
            \setlength{\figureheight}{1.925\textwidth}
            \setlength{\figurewidth}{1.23\textwidth}
            \scriptsize{
\begin{tikzpicture}

\definecolor{color0}{rgb}{1,0.498039215686275,0.0549019607843137}

\begin{axis}[
height=\figureheight,
tick align=outside,
width=\figurewidth,
x grid style={lightgray!92.02614379084967!black},
xmin=0.5, xmax=1.5,
xtick={1},
xticklabels={Distribution},
y grid style={lightgray!92.02614379084967!black},
ytick style={color=black},
]
\addplot [black]
table {%
0.925 -0.09890431762
1.075 -0.09890431762
1.075 0.49437728107075
0.925 0.49437728107075
0.925 -0.09890431762
};
\addplot [black]
table {%
1 -0.09890431762
1 -0.98668132782
};
\addplot [black]
table {%
1 0.49437728107075
1 1.37918371201
};
\addplot [black]
table {%
0.9625 -0.98668132782
1.0375 -0.98668132782
};
\addplot [black]
table {%
0.9625 1.37918371201
1.0375 1.37918371201
};
\addplot [black, mark=*, mark size=0.42, mark options={solid,fill opacity=0}, only marks]
table {%
1 -0.99158584595
1 -1.17894368172
1 -1.24991952896
1 -1.2677794838
1 -1.04723946571
1 -1.09644481659
1 -1.20672019005
1 -1.3064193058
1 -0.99583470345
1 -1.17657475471
1 -0.99652230263
1 -1.12641572952
1 -1.21645220757
1 -1.0760933876
1 -0.99617611885
1 -1.14599182129
1 -1.31639021873
1 -1.6364538765
1 -1.60638878822
1 -1.54651491165
1 -1.0163666153
1 -1.16652659416
1 -1.28646985054
1 -0.99623095512
1 -1.03855744362
1 -1.0162290287
1 -1.16761352539
1 -0.98988493919
1 -1.30897823334
1 -1.46786243439
1 -1.5682360363
1 -1.6288096714
1 -1.64040159225
1 -1.63040159225
1 -1.36523076057
1 -1.12402603149
1 -0.99435910225
1 -1.04827004433
1 -1.05114758492
1 -1.06453241348
1 -1.07659664154
1 -1.00803256989
1 -1.03104762077
1 -1.13942317963
1 -1.48859883308
1 -1.10486024857
1 -1.54398052216
1 -1.34609892845
1 -1.38633126259
1 -1.16398396492
1 4.69805517197
1 3.64183670998
1 2.88882297516
1 2.23002380371
1 1.74281734467
1 1.45499738693
1 1.50183380127
1 3.24771688461
1 2.55462748528
1 1.93165967941
1 1.44165967941
1 1.51146715641
1 2.18317355633
1 2.77191761494
1 1.81224941254
1 2.490863560438
1 1.871367875338
1 1.394818100929
1 1.48621900558
1 1.61147593498
1 1.94362861633
1 1.45369227409
1 1.45232115269
1 2.87857952595
1 3.73857952595
1 3.53093698978
1 3.21860841751
1 2.67636293411
1 2.097744224072
1 1.583151757717
1 1.595493381023
1 1.42057818532
1 4.53712294579
1 3.07071466446
1 2.37960456848
1 1.85960456848
1 2.44780328751
1 5.20568632126
1 5.90013353348
1 6.91556791306
1 6.39503257751
1 5.05100606918
1 4.09947559357
1 3.4228191185
1 2.91158384323
1 2.48237060547
1 1.65308179855
1 2.7460247612
1 1.84068035126
1 4.00040599823
1 3.02885206223
1 2.41885206223
1 1.86244232178
1 1.49575304031
1 1.77284100533
1 1.57161863804
1 1.43459684849
1 1.50313796997
1 1.51233945847
1 2.32610786438
1 1.67958971024
1 1.7721137619
1 1.47731431961
1 1.57720754623
1 1.85641073227
1 1.58475460052
1 1.49525167465
1 2.31628656387
1 3.29500482559
1 3.74628608704
1 3.67181192398
1 3.55174325943
1 3.69214857101
1 3.65286668777
1 2.71286668777
1 2.26210586548
1 2.52390138626
1 1.97429069519
1 1.80185857773
1 2.77405822754
1 2.32755439758
1 2.18482736588
1 1.85449884415
1 1.43329391479
1 1.47136537552
1 3.75483595848
1 3.09320706367
1 2.38919208527
1 2.05906646729
1 1.40985954285
1 1.40068431854
1 4.06915359497
1 3.51915359497
1 3.30165712357
1 2.7469793129
1 1.9269793129
1 1.75928762436
1 1.95928762436
1 2.22326135635
1 1.84240203857
1 1.7318655014
1 2.5827362442
1 1.57133989334
};
\addplot [color0]
table {%
0.925 0.260524082185
1.075 0.260524082185
};
\end{axis}

\end{tikzpicture}}
        \end{minipage}
        \centering
        \hspace{0.42cm} \scriptsize{(a) X-coordinate error}
    \end{minipage}
    \begin{minipage}{0.24\textwidth}
        \begin{minipage}{0.56\textwidth}
            \centering
            \setlength{\figureheight}{1.375\textwidth}
            \setlength{\figurewidth}{1.23\textwidth}
            \scriptsize{\input{fig/one_corner/y_error.tex}} \\
        \end{minipage}\hspace{-0.12cm}
        \begin{minipage}{0.44\textwidth}
            \centering
            \setlength{\figureheight}{1.925\textwidth}
            \setlength{\figurewidth}{1.23\textwidth}
            \scriptsize{
\begin{tikzpicture}

\definecolor{color0}{rgb}{1,0.498039215686275,0.0549019607843137}

\begin{axis}[
height=\figureheight,
tick align=outside,
width=\figurewidth,
x grid style={lightgray!92.02614379084967!black},
xmin=0.5, xmax=1.5,
xtick={1},
xticklabels={Distribution},
y grid style={lightgray!92.02614379084967!black},
ytick style={color=black},
]
\addplot [black]
table {%
0.925 -0.5370318382965
1.075 -0.5370318382965
1.075 0.21157142282
0.925 0.21157142282
0.925 -0.5370318382965
};
\addplot [black]
table {%
1 -0.5370318382965
1 -1.6475190132856
};
\addplot [black]
table {%
1 0.21157142282
1 1.31601451874
};
\addplot [black]
table {%
0.9625 -1.6475190132856
1.0375 -1.6475190132856
};
\addplot [black]
table {%
0.9625 1.31601451874
1.0375 1.31601451874
};
\addplot [black, mark=*, mark size=0.42, mark options={solid,fill opacity=0}, only marks]
table {%
1 -2.15689738274
1 -2.45190952778
1 -1.78190952778
1 -1.91139906406
1 -2.47164499283
1 -1.82164499283
1 -2.42302090645
1 -3.03302090645
1 -3.49302090645
1 -2.6757231617
1 -1.95984863758
1 -1.91051199970767
1 -2.39444844760001
1 -3.73391711130738
1 -2.94383381567895
1 -2.36339138437063
1 -2.13339138437063
1 -1.76339138437063
1 -2.37395330436528
1 -1.73392418842763
1 -3.27420125037432
1 -2.62417561292648
1 -2.14374471604824
1 -1.78385794438422
1 -1.98423213116825
1 -2.89385297019035
1 -2.37376962706447
1 -1.72425690807402
1 -2.63418440461159
1 -1.91370526149869
1 -1.77374476354569
1 -2.11380921851844
1 -1.94370535090566
1 -2.03425924941897
1 -1.87334773048759
1 -2.28449422210455
1 -3.68449422210455
1 -3.08449422210455
1 -4.39449422210455
1 -3.26449422210455
1 -4.47449422210455
1 -3.47449422210455
1 -2.69449422210455
1 -2.32449422210455
1 -1.93429106339812
1 -2.72386662989855
1 -2.20414868839085
1 -3.62389833584428
1 -2.66418058805168
1 -2.28372425861657
1 -1.67460451584309
1 -2.6988762962073
1 -2.1336473520845
1 -1.6618972423673
1 -1.89862664223
1 -2.54812790394
1 -2.50352200508
1 -2.32346563816
1 -2.68751560688
1 -3.5884318924
1 -3.76760237694
1 -4.0612116003
1 -2.96695302963
1 -2.95929400444
1 -2.54257766724
1 -2.29702364922
1 -1.82625132561
1 -1.81680847168
1 -2.23055012703
1 -5.86488883018
1 -6.80655963898
1 -9.22342347145
1 -7.8622299099
1 -6.92765392303
1 -6.36181217194
1 -6.066379776
1 -4.24402982712
1 -2.86402982712
1 -2.13179879189
1 -3.95987186432
1 -3.0612806797
1 -2.63744880676
1 -2.20942552567
1 -2.222811546326
1 -3.56953088522
1 -4.56979296207
1 -3.42142551422
1 -2.61778871059
1 -3.37390468597
1 -2.90599374294
1 -3.54638252258
1 -5.412076849937
1 -6.798964960575
1 -7.617154985666
1 -5.349837548137
1 -4.1665827184916
1 -4.281413234472
1 -6.041501142979
1 -3.895444777012
1 -3.966896491051
1 -3.082039105892
1 -2.11738790035
1 -2.00301199913
1 -2.90908792496
1 -3.02872816563
1 -2.26000772476
1 -3.15074329376
1 -3.09479430199
1 -3.47461361885
1 -3.78300467968
1 -3.37126423359
1 -2.63974636078
1 -2.06881428242
1 -2.44842599869
1 -2.20974700451
1 -2.57087146282
1 -3.17269721985
1 -3.06742794037
1 -2.35044589043
1 -1.709685538113
1 -1.873393278122
1 -2.116251683235
1 -1.814440908432
1 -2.076418266296
1 -1.910823397636
1 -2.11387878418
1 -2.01387878418
1 -1.67387878418
1 -1.90387878418
1 -4.43544775963
1 -3.10248903275
1 -2.46355360985
1 -1.66508024216
1 -4.70326517105
1 -3.13326517105
1 -2.02066086769
1 1.40358954072
1 1.457402011156
1 1.5061548545957
1 1.41614719744772
1 1.495864295959
1 1.553919908404
1 1.595653002858
1 1.553205502033
1 1.339231752753
1 1.87683211327
1 2.64954472065
1 3.09016600132
1 2.59211540222
1 2.19945128441
1 1.59020373344
1 1.5604020071
1 1.73137788296
1 2.88202888489
1 3.3566299057
1 1.75241679192
1 2.40843637466
1 1.53770537376
1 1.67692194939
1 1.38097100258
};
\addplot [color0]
table {%
0.925 -0.127899650039
1.075 -0.127899650039
};
\end{axis}

\end{tikzpicture}}
        \end{minipage}
        \centering
        \hspace{0.42cm} \scriptsize{(b) Y-coordinate error}
    \end{minipage}
    \caption{Localization error when all the four anchors are nearby, defining three faces of a cube. While the average error is small, and 50\,\% of the measurements are relatively accurate, the localization estimation is highly unstable. Further filtering is needed to enable accurate flight in this case.}
    \label{fig:one_corner_anchors}
\end{figure}

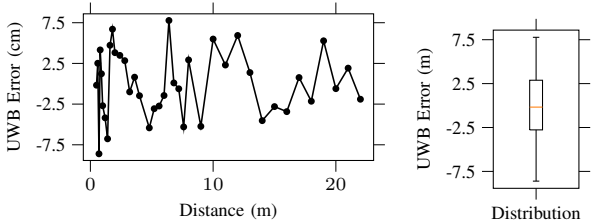
\begin{figure}
    \centering
            \setlength{\figureheight}{0.1948\textwidth}
            \setlength{\figurewidth}{0.3\textwidth}
            \scriptsize{
\begin{tikzpicture}

\begin{axis}[
height=\figureheight,
tick align=outside,
width=\figurewidth,
x grid style={white!69.01960784313725!black},
xlabel={Distance (m)},
scaled y ticks = false,
xmin=-0.575, xmax=23.075,
xtick style={color=black},
y grid style={white!69.01960784313725!black},
ylabel={UWB Error (cm)},
ymin=-0.0942754199921553, ymax=0.0859319778000276,
ytick style={color=black},
ytick={-0.075,-0.025,0.025,0.075},
yticklabels={-7.5, -2.5, 2.5, 7.5}
]
\addplot [semithick, black, mark=*, mark size=1, mark options={solid}]
table {%
0.5 -0.00177118537506026
0.6 0.0251827276349348
0.7 -0.0860841746379651
0.8 0.0415751557384223
0.9 0.0122442533608242
1 -0.0270215421323087
1.2 -0.041882207043989
1.4 -0.0676689175935039
1.6 0.0472711632970291
1.8 0.0670452492085366
2 0.0382015898357113
2.4 0.0348643600489845
2.8 0.0283878637878382
3.2 -0.0100031728982822
3.6 0.00808729963108362
4 -0.0146916042447202
4.8 -0.0542119969903325
5.2 -0.0306401866125293
5.6 -0.0273468388131914
6 -0.0144403978480193
6.4 0.0777407324458375
6.8 0.00101903555537774
7.2 -0.00618233633843602
7.6 -0.0530356709492605
8 0.0292751446684497
9 -0.0523509328012096
10 0.0549226772031909
11 0.0230124288407706
12 0.0593345496024335
13 0.013748118772894
14 -0.0453219866914719
15 -0.0280666357977179
16 -0.0342498847380489
17 0.0076300612083196
18 -0.0214845279665974
19 0.0527587199821937
20 -0.00601094675604372
21 0.0191868641826932
22 -0.0190228487769264
};
\end{axis}

\end{tikzpicture}}
            \setlength{\figureheight}{0.2033\textwidth}
            \setlength{\figurewidth}{0.15\textwidth}
            \scriptsize{
\begin{tikzpicture}

\definecolor{color0}{rgb}{1,0.498039215686275,0.0549019607843137}

\begin{axis}[
height=\figureheight,
width=\figurewidth,
tick align=outside,
width=\figurewidth,
x grid style={lightgray!92.02614379084967!black},
xmin=0.5, xmax=1.5,
xtick style={color=black},
xtick={1},
xticklabels={Distribution},
x grid style={white!69.01960784313725!black},
xmin=0.5, xmax=1.5,
y grid style={white!69.01960784313725!black},
ymin=-0.0942754199921553, ymax=0.0859319778000276,
ytick style={color=black},
scaled y ticks = false,
ytick={-0.075,-0.025,0.025,0.075},
yticklabels={-7.5,-2.5,2.5,7.5},
ylabel={UWB Error (m)},
]
\addplot [black]
table {%
0.925 -0.0277067373054547
1.075 -0.0277067373054547
1.075 0.028831504228144
0.925 0.028831504228144
0.925 -0.0277067373054547
};
\addplot [black]
table {%
1 -0.0277067373054547
1 -0.0860841746379651
};
\addplot [black]
table {%
1 0.028831504228144
1 0.0777407324458375
};
\addplot [black]
table {%
0.9625 -0.0860841746379651
1.0375 -0.0860841746379651
};
\addplot [black]
table {%
0.9625 0.0777407324458375
1.0375 0.0777407324458375
};
\addplot [color0]
table {%
0.925 -0.00177118537506026
1.075 -0.00177118537506026
};
\end{axis}

\end{tikzpicture}}
    \caption{Average distance estimation error between two DWM1001 nodes in line of sight. The nodes have been calibrated to take into account the antenna delay. The mean error is smaller than 1~mm while the standard deviation is 3.9\,cm. The maximum error is 8.6\,cm.}
    \label{fig:one2one_errors}
\end{figure}

\begin{figure}[t]
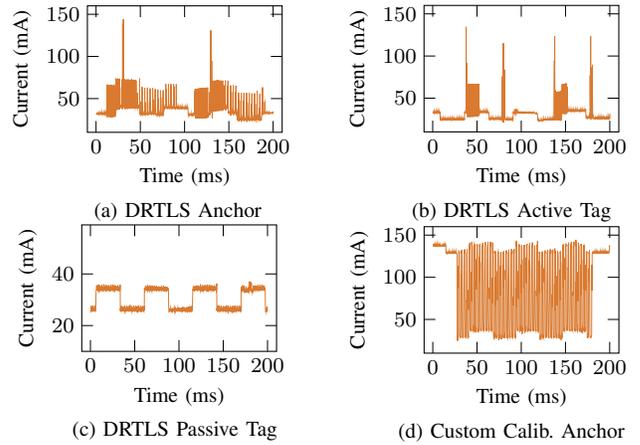

    \centering
    \setlength{\figurewidth}{0.23\textwidth}
    \setlength{\figureheight}{0.18\textwidth}
    \begin{minipage}{0.24\textwidth}
        \centering
        \footnotesize{\input{fig/uwb_current/Anchor_no_initiator_all.tex}} \\
        \hspace{0.6cm} (a) DRTLS Anchor
    \end{minipage}
    \begin{minipage}{0.24\textwidth}
        \centering
        \footnotesize{\input{fig/uwb_current/tag_active_all.tex}} \\
        \hspace{0.52cm} (b) DRTLS Active Tag
    \end{minipage}
    
    $ $ \\
    \begin{minipage}{0.24\textwidth}
        \centering
        \footnotesize{\input{fig/uwb_current/tag_pasive_all.tex}} \\
        \hspace{0.52cm} (c) DRTLS Passive Tag
    \end{minipage}
    \begin{minipage}{0.24\textwidth}
        \centering
        \footnotesize{\input{fig/uwb_current/initiator_all_samples.tex}} \\
        \hspace{0.52cm} (d) Custom Calib. Anchor
    \end{minipage}

    \caption{Power consumption of the UWB anchors and tags in different modes. The tags were powered through USB in all these measurements. In the self-reconfigurable anchor, the communication is constant while calculating and therefore the power consumption is greater than in Decawave's DRTL setup.}
    \label{fig:power}
\end{figure}

\section{Experimentation and Results}

To test the feasibility of an autonomous flight based only on UWB ToF measurements and a 1D lidar for height estimation, we show a simple circular trajectory. 

\subsection{Flight with UWB-based localization}

Figure~\ref{fig:circle} shows the path and localization error recorded over an autonomous flight using UWB for localization in the horizontal plane and a 1D lidar for height estimation. The control input, in green in the figure, given to the drone in the form of waypoints is described by \eqref{eq:control_circle}:

\begin{dmath}
    \textbf{p}_{obj} = \left(r_{obj}(t), \theta_{obj}(t), z_{obj}(t)\right) = 
    \left\{
    \begin{array}{ll}
        \left(r_0, \theta_0, z_0\right) & if \: \left\vert \left\lVert p(t) \right\lVert - \left\lVert p_0 \right\rVert \right\vert  > \varepsilon \\[+5pt]
        \left(r_0, \theta(t) + \Delta\theta, z_0 \right) & otherwise
    \end{array}
    \right.
    \label{eq:control_circle}
\end{dmath}

\noindent where $\textbf{p}(t) = \left(r(t), \theta(t), z(t)\right)$ represents the current position of the UAV, $\textbf{p}_{obj}$ is the waypoint given to the UAV as its objective position, $p_0 = \left(r_0, \theta_0, z_0 \right)$ is the entry position to the circular path, $\varepsilon$ is a threshold to consider that the UAV is within the predefined path, and $\Delta\theta$ defines the angular speed when it is considered together with the waypoint update rate and the maximum linear speed of the UAV. In the experiment shown in Figure~\ref{fig:circle}, we have defined $p_0=(1.23,0,1.23)$, $\varepsilon = 0.3\,m$ and $\Delta\theta = 0.05\,rad$. All the positions are represented in cylindrical coordinates, and the radial distance and height are given in meters.

\subsection{Spatial distribution of anchors}

One of the novel analysis in this paper is the study of different spatial anchor distributions. The accuracy recorded during the autonomous drone flight in Fig.~\ref{fig:circle}\,(d)-(f) relates to a typical setting where the anchors are positioned in the corners of a room. One thing to note in this case, nonetheless, is the enhanced accuracy of the latest DWM1001 transceiver. Through our experiments, we have also noted that over 50\,\% of the samples acquired during the flight had an error under 10\,cm. In particular, the lower and upper quartiles in the case of the x-coordinate error in Fig.~\ref{fig:circle}\,(d) reflect an error smaller than 5\,cm.

An extreme case for the location of anchors would be when these are located in a single robot or movable unit. This has been done, for instance, by Nguyen \textit{et al.} in~\cite{nguyen2019integrated} for an UAV to dock on a moving platform. However, in that article, the authors report using a movable platform of $3\,m^2$, which is impractical in most situations, in particular in post-disaster scenarios where access to the objective operation area might be limited. In order to assess the viability of a more practical usage, we have located four anchors in four corners of a cube, with one of them representing the origin of coordinates and each of the other three the axes. The separation between the anchor in the origin and each of the others was just 60\,cm. Fig.~\ref{fig:one_corner_anchors} shows the position estimation error in this case. We can see that the position estimation is highly unstable. However, the lower and upper quartile in the boxplot relate to relatively low errors. With proper filtering and sensor fusion it might be possible to utilize such anchor settings when the error margin allows.

\begin{figure*}
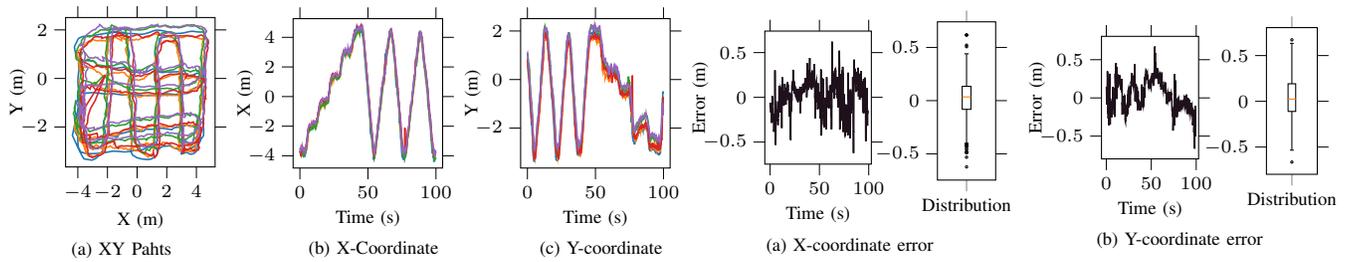

    \centering

    \begin{minipage}{0.16\textwidth}
        \centering
        \setlength{\figureheight}{1.23\textwidth}
        \setlength{\figurewidth}{1.23\textwidth}
        \scriptsize{\input{fig/anchors_in_center/xy_paths.tex}} \\
        \hspace{0.15cm} (a) XY Pahts
    \end{minipage}
    \begin{minipage}{0.16\textwidth}
        \centering
        \setlength{\figureheight}{1.23\textwidth}
        \setlength{\figurewidth}{1.23\textwidth}
        \scriptsize{\input{fig/anchors_in_center/x_trajectory.tex}} \\
        \hspace{0.8cm} (b) X-Coordinate
    \end{minipage}
    \begin{minipage}{0.16\textwidth}
        \centering
        \setlength{\figureheight}{1.23\textwidth}
        \setlength{\figurewidth}{1.23\textwidth}
        \scriptsize{\input{fig/anchors_in_center/y_trajectory.tex}} \\
        \hspace{0.8cm} (c) Y-coordinate
    \end{minipage}
    \begin{minipage}{0.24\textwidth}
        \begin{minipage}{0.56\textwidth}
            \centering
            \setlength{\figureheight}{1.375\textwidth}
            \setlength{\figurewidth}{1.23\textwidth}
            \scriptsize{\input{fig/anchors_in_center/x_error.tex}} \\
        \end{minipage}\hspace{-0.12cm}
        \begin{minipage}{0.44\textwidth}
            \centering
            \setlength{\figureheight}{1.925\textwidth}
            \setlength{\figurewidth}{1.23\textwidth}
            \scriptsize{
\begin{tikzpicture}

\definecolor{color0}{rgb}{1,0.498039215686275,0.0549019607843137}

\begin{axis}[
height=\figureheight,
tick align=outside,
width=\figurewidth,
x grid style={lightgray!92.02614379084967!black},
xmin=0.5, xmax=1.5,
xtick={1},
xticklabels={Distribution},
y grid style={lightgray!92.02614379084967!black},
ytick style={color=black},
]

\addplot [black]
table {%
0.925 -0.081722430586
1.075 -0.081722430586
1.075 0.134642832275
0.925 0.134642832275
0.925 -0.081722430586
};
\addplot [black]
table {%
1 -0.081722430586
1 -0.40270938873
};
\addplot [black]
table {%
1 0.134642832275
1 0.43999191284
};
\addplot [black]
table {%
0.9625 -0.40270938873
1.0375 -0.40270938873
};
\addplot [black]
table {%
0.9625 0.43999191284
1.0375 0.43999191284
};
\addplot [black, mark=*, mark size=0.42, mark options={solid,fill opacity=0}, only marks]
table {%
1 -0.53174873352
1 -0.48360512733
1 -0.49012638092
1 -0.40700314045
1 -0.47700314045
1 -0.42802527428
1 -0.425254535680001
1 -0.43051643372
1 -0.4273440361
1 -0.4242293644
1 -0.45541227818
1 -0.41275700569
1 -0.45262408733
1 -0.62262408733
1 -0.4907434082
1 0.61659622192
1 0.61659622192
1 0.61659622192
1 0.50659622192
1 0.52125956535
};
\addplot [color0]
table {%
0.925 0.03523575306
1.075 0.03523575306
};
\end{axis}

\end{tikzpicture}}
        \end{minipage}
        \centering
        \hspace{0.23cm} \scriptsize{(a) X-coordinate error}
    \end{minipage}
    \begin{minipage}{0.23\textwidth}
        \begin{minipage}{0.56\textwidth}
            \centering
            \setlength{\figureheight}{1.375\textwidth}
            \setlength{\figurewidth}{1.23\textwidth}
            \scriptsize{\input{fig/anchors_in_center/y_error.tex}} \\
        \end{minipage}\hspace{-0.12cm}
        \begin{minipage}{0.44\textwidth}
            \centering
            \setlength{\figureheight}{1.925\textwidth}
            \setlength{\figurewidth}{1.23\textwidth}
            \scriptsize{
\begin{tikzpicture}

\definecolor{color0}{rgb}{1,0.498039215686275,0.0549019607843137}

\begin{axis}[
height=\figureheight,
tick align=outside,
width=\figurewidth,
x grid style={lightgray!92.02614379084967!black},
xmin=0.5, xmax=1.5,
xtick={1},
xticklabels={Distribution},
y grid style={lightgray!92.02614379084967!black},
ytick style={color=black},
]
\addplot [black]
table {%
0.925 -0.1104566788675
1.075 -0.1104566788675
1.075 0.1917641985425
0.925 0.1917641985425
0.925 -0.1104566788675
};
\addplot [black]
table {%
1 -0.1104566788675
1 -0.53422292709
};
\addplot [black]
table {%
1 0.1917641985425
1 0.63381429672
};
\addplot [black]
table {%
0.9625 -0.53422292709
1.0375 -0.53422292709
};
\addplot [black]
table {%
0.9625 0.63381429672
1.0375 0.63381429672
};
\addplot [black, mark=*, mark size=0.42, mark options={solid,fill opacity=0}, only marks]
table {%
1 -0.66422292709
1 0.67214551449
};
\addplot [color0]
table {%
0.925 0.024590930939
1.075 0.024590930939
};
\end{axis}

\end{tikzpicture}}
        \end{minipage}
        \centering
        \hspace{0.23cm} \scriptsize{(b) Y-coordinate error}
    \end{minipage}
    \caption{Data recorded with 4 tags (single rigid body) and four anchors positioned in the center of the motion capture arena, separated 1.8~m only. The legend is not included due to lack of space. Blue represents the Optitrack ground truth, while the other four colors each represent one of the tags.}
    \label{fig:anchors_center}
\end{figure*}

The last anchor distribution included in the dataset represents, to the best of our knowledge, the most usable for a moving platform, with four anchors near the ground and separated only around 2\,m. One of the flights recorded and the corresponding errors are illustrated in Fig.~\ref{fig:anchors_center}. The localization accuracy in this case doubles but still allows the possibility of autonomous flight.

\subsection{Autopositioning of anchors}

The localization estimation provided by Decawave's UART API has given us better results than the utilization of raw individual distance measurements applied to multiple open-source multilateration algorithms. The code to interface the API with ROS for both passive and active tags is made available, together with the data, in the Github repository. However, Decawave's function to autoposition the anchors has not given good results in our experience. Moreover, the calculation takes around 40\,s, which is unassumable in some mobile settings. In order to tackle this issue, we have written our own firmware for the anchors in order to recalculate their position if they move. In our experiments, we utilize separate UWB devices for the autopositioning. Each anchor location is equipped with one device flashed with our code for autopositioning only, and another one as an anchor for the localization of the UWB tag, flashed with Decawave's proprietary firmware. We utilize the UART API to set the anchor positions after the autopositioning. In a real scenario, a single device could be used as both but it would need to be reprogrammed on the fly.

Figure~\ref{fig:one2one_errors} shows the average error of the distance estimation between two anchors during the self-calibration process. The measurements are taken at over 35 different distances up to 22\,m, with 50 measurements for each distance. The standard deviation for each particular distance ranged from 0 to 4\,cm, while the standard deviation of the error altogether was under 3\,cm in over 50\,\% of the cases, as shown in the boxplot in Fig.~\ref{fig:one2one_errors}. The comparison between Decawave's autopositioning and ours is shown in Table~\ref{tab:autopositioning}.

In addition, we have measured the power consumption of the UWB devices in different modes, as shown in Table~\ref{tab:power} and Fig.~\ref{fig:power}. This includes anchors, active tags and passive tags, as well as the device running our autopositioning firmware. In the latter case, we provide an initial implementation with no power usage optimization, and during the autopositioning the nodes are transmitting at high frequency. Therefore the power consumption is very high. We differentiate between \textit{responder} and \textit{initiator} types. Each of the four anchors takes the role of \textit{initiator} one time, sending a message one by one to each of the other three (in \textit{responder} mode), and calculating the distance via two-way ranging. The distance between two nodes is thus calculated twice during the autopositioning.

During the autopositioning process, the first anchor to become an initiator, which is activated via a \textit{start} command through the UART interface, is considered the origin of coordinates. Then, we assume that some information about the position of the other anchors exists. The minimum information required is to know the order of the anchors over the boundary of their convex envelope in a counter-clockwise direction. We also predefine the \textit{x} axis to follow the direction of the vector that is defined from the first to the second anchor following the aforementioned order.

\subsection{Characterization of UWB localization accuracy}

We have classified the accuracy of the UWB localization based on the distance to the center of mass of the anchor system, the height of the UAV and its speed.

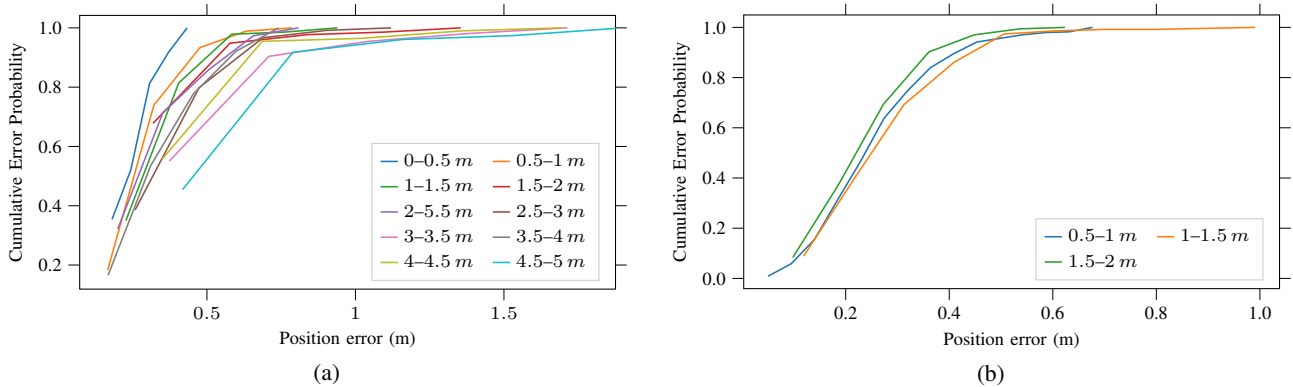
\begin{figure*}
    \centering
    \begin{minipage}{0.48\textwidth}
        \centering
        \setlength{\figureheight}{0.6\textwidth}
        \setlength{\figurewidth}{\textwidth}
        \scriptsize{
\begin{tikzpicture}

\definecolor{color6}{rgb}{0.890196078431372,0.466666666666667,0.76078431372549}
\definecolor{color8}{rgb}{0.0901960784313725,0.745098039215686,0.811764705882353}
\definecolor{color7}{rgb}{0.737254901960784,0.741176470588235,0.133333333333333}
\definecolor{color2}{rgb}{0.172549019607843,0.627450980392157,0.172549019607843}
\definecolor{color4}{rgb}{0.580392156862745,0.403921568627451,0.741176470588235}
\definecolor{color3}{rgb}{0.83921568627451,0.152941176470588,0.156862745098039}
\definecolor{color5}{rgb}{0.549019607843137,0.337254901960784,0.294117647058824}
\definecolor{color1}{rgb}{1,0.498039215686275,0.0549019607843137}
\definecolor{color0}{rgb}{0.12156862745098,0.466666666666667,0.705882352941177}


\begin{axis}[
    height=\figureheight,
    legend cell align={left},
    legend style={fill opacity=0.8, draw opacity=1, text opacity=1, draw=white!80.0!black}, 
    legend pos=south east,
    tick align=outside,
    width=\figurewidth,
    x grid style={white!69.01960784313725!black},
    xlabel={Position error (m)},
    xmin=0.0715837654642694, xmax=1.87561393713688,
    xtick style={color=black},
    y grid style={white!69.01960784313725!black},
    ylabel={Cumulative Error Probability},
    ymin=0.11875, ymax=1.04196428571429,
    ytick style={color=black},
    ytick={0,0.2,0.4,0.6,0.8,1,1.2},
    yticklabels={0.0,0.2,0.4,0.6,0.8,1.0,1.2},
    legend columns=2, 
    /tikz/column 2/.style={
        column sep=5pt,
    },
    legend image code/.code={
        \draw[mark repeat=2,mark phase=2]
        plot coordinates {
            (0cm,0cm)
            (0.1cm,0cm) 
            (0.23cm,0cm)  
        };%
    },
]
\addplot [semithick, color0]
table {%
0.180525208346629 0.354166666666667
0.243729596140822 0.520833333333333
0.306933983935015 0.8125
0.370138371729208 0.916666666666666
0.433342759523401 1
};
\addlegendentry{$\numrange{0}{0.5}\,m$}
\addplot [semithick, color1]
table {%
0.165944515192492 0.183333333333333
0.321255970277667 0.738888888888889
0.476567425362843 0.933333333333333
0.631878880448018 0.988888888888889
0.787190335533194 1
};
\addlegendentry{$\numrange{0.5}{1}\,m$}
\addplot [semithick, color2]
table {%
0.227099782750357 0.35
0.405190130271079 0.814285714285714
0.583280477791802 0.978571428571428
0.761370825312525 0.985714285714286
0.939461172833247 1
};
\addlegendentry{$\numrange{1}{1.5}\,m$}
\addplot [semithick, color3]
table {%
0.3187255346395 0.67816091954023
0.577718402548949 0.948275862068965
0.836711270458399 0.977011494252873
1.09570413836785 0.985632183908046
1.3546970062773 1
};
\addlegendentry{$\numrange{1.5}{2}\,m$}
\addplot [semithick, color4]
table {%
0.199698035075374 0.321428571428571
0.351913905626975 0.714285714285714
0.504129776178577 0.857142857142857
0.656345646730178 0.973214285714285
0.808561517281779 1
};
\addlegendentry{$\numrange{2}{5.5}\,m$}
\addplot [semithick, color5]
table {%
0.257977661087019 0.385416666666667
0.473411131011377 0.797916666666667
0.688844600935734 0.966666666666667
0.904278070860092 0.991666666666667
1.11971154078445 1
};
\addlegendentry{$\numrange{2.5}{3}\,m$}
\addplot [semithick, color6]
table {%
0.373388995840862 0.550699300699301
0.708414408569125 0.903846153846154
1.04343982129739 0.954545454545455
1.37846523402565 0.980769230769231
1.71349064675391 1
};
\addlegendentry{$\numrange{3}{3.5}\,m$}
\addplot [semithick, gray!99.6078431372549!black]
table {%
0.167201849736644 0.165354330708661
0.310809709557114 0.535433070866142
0.454417569377583 0.777559055118111
0.598025429198052 0.921259842519685
0.741633289018521 1
};
\addlegendentry{$\numrange{3.5}{4}\,m$}
\addplot [semithick, color7]
table {%
0.350149233155196 0.558080808080808
0.686123924900025 0.954545454545454
1.02209861664485 0.964646464646464
1.35807330838968 0.98989898989899
1.69404800013451 1
};
\addlegendentry{$\numrange{4}{4.5}\,m$}
\addplot [semithick, color8]
table {%
0.418003299690483 0.454347826086957
0.788960765654373 0.917391304347826
1.15991823161826 0.960869565217391
1.53087569758215 0.973913043478261
1.90183316354604 1
};
\addlegendentry{$\numrange{4.5}{5}\,m$}
\end{axis}

\end{tikzpicture}} \\
        \hspace{0.23cm} \small{(a)}
    \end{minipage}
    \begin{minipage}{0.48\textwidth}
        \centering
        \setlength{\figureheight}{0.6\textwidth}
        \setlength{\figurewidth}{\textwidth}
        \scriptsize{
\begin{tikzpicture}

\definecolor{color1}{rgb}{1,0.498039215686275,0.0549019607843137}
\definecolor{color0}{rgb}{0.12156862745098,0.466666666666667,0.705882352941177}
\definecolor{color2}{rgb}{0.172549019607843,0.627450980392157,0.172549019607843}

\begin{axis}[
    height=\figureheight,
    legend cell align={left},
    legend style={fill opacity=0.8, draw opacity=1, text opacity=1, draw=white!80.0!black}, 
    legend pos=south east,
    tick align=outside,
    width=\figurewidth,
    x grid style={lightgray!92.02614379084967!black},
    xlabel={Position error (m)},
    xmin=0.00303424543339768, xmax=1.0379244137707,
    xtick style={color=black},
    xtick={0,0.2,0.4,0.6,0.8,1,1.2},
    xticklabels={0.0,0.2,0.4,0.6,0.8,1.0,1.2},
    y grid style={lightgray!92.02614379084967!black},
    ylabel={Cumulative Error Probability},
    ymin=-0.0407894736842105, ymax=1.04956140350877,
    ytick style={color=black},
    ytick={-0.2,0,0.2,0.4,0.6,0.8,1,1.2},
    yticklabels={−0.2,0.0,0.2,0.4,0.6,0.8,1.0,1.2},
    legend columns=2, 
    /tikz/column 2/.style={
        column sep=5pt,
    },
    legend image code/.code={
        \draw[mark repeat=2,mark phase=2]
        plot coordinates {
            (0cm,0cm)
            (0.1cm,0cm) 
            (0.23cm,0cm)  
        };%
    },
]
\addplot [semithick, color0]
table {%
0.0500747076305477 0.0087719298245614
0.0948371041650188 0.0584795321637427
0.13959950069949 0.154970760233918
0.184361897233961 0.312865497076023
0.229124293768432 0.467836257309942
0.273886690302903 0.637426900584795
0.318649086837375 0.745614035087719
0.363411483371846 0.839181286549708
0.408173879906317 0.894736842105263
0.452936276440788 0.941520467836257
0.497698672975259 0.956140350877193
0.54246106950973 0.970760233918128
0.587223466044202 0.97953216374269
0.631985862578673 0.982456140350877
0.676748259113144 1
};
\addlegendentry{$\numrange{0.5}{1}\,m$}
\addplot [semithick, color1]
table {%
0.11862790104269 0.0885311871227364
0.215545239990563 0.394366197183099
0.312462578938436 0.69215291750503
0.409379917886309 0.861167002012072
0.506297256834182 0.973843058350101
0.603214595782055 0.985915492957746
0.700131934729928 0.99195171026157
0.797049273677801 0.99195171026157
0.893966612625674 0.995975855130785
0.990883951573547 1
};
\addlegendentry{$\numrange{1}{1.5}\,m$}
\addplot [semithick, color2]
table {%
0.0977862110788863 0.0830860534124629
0.185407287754292 0.370919881305638
0.273028364429698 0.694362017804154
0.360649441105104 0.902077151335312
0.44827051778051 0.970326409495549
0.535891594455916 0.99406528189911
0.623512671131322 1
};
\addlegendentry{$\numrange{1.5}{2}\,m$}
\end{axis}

\end{tikzpicture}} \\
        \hspace{0.23cm} \small{(b)}
    \end{minipage}
    \caption{Positioning error cumulative probability for different distances to the center of mass of the anchor system (a), and different heights (b). The anchor positions were $(1.79, 0.58, 0.1)$, $(0.01, 0.58, 0.1)$, $(1.8, -1.2, 0.1)$, and $(0,-1.21, 0.1)$.}
    \label{fig:center_height_errors}
\end{figure*}

The dataset introduced in this paper includes data of an UAV flying both inside and outside the convex envelope defined by the anchor positions. In general, the closer an UWB tag is to the center of mass of the anchor system, the higher the position estimation accuracy is. This is illustrated in Fig.~\ref{fig:corners_cumulatives}\,(a) for the case where the anchors are in the corners of the room, and in Fig.~\ref{fig:center_height_errors}\,(a) for the case in which the anchors are in the center and close to the floor.

Based on the rest of the measurements in the same two figures, we can also see that the position estimation error is smaller with lower speed, as illustrated in Fig.~\ref{fig:corners_cumulatives}\,(c). Regarding the height, when the anchors were all near the floor, we did not obtain significant differences, as sown in Fig.~\ref{fig:center_height_errors}\,(b). However, when the anchors were at the height of 1.8\,m in the corners of the room, higher flight altitudes resulted in smaller errors, as Fig.~\ref{fig:corners_cumulatives}\,(b) shows. While LOS was always ensured during the experiments, the error was smallest near the constant \textit{z} plane defined by the anchor positions.

The conclusions from the above characterization of accuracy based on speed, height and distance to the center of mass of the anchor system allow a more efficient control of autonomous UAVs in a real deployment, where strategies can be defined to adjust the error estimation based on these parameters.

\begin{table}
    \centering
    \caption{Power conssumption of UWB tags and anchors.}
    \label{tab:power}
    \begin{tabular}{@{}lcccc@{}}
        \toprule
                & \multicolumn{2}{c}{Power @ 5V} & \multicolumn{2}{c}{Power @ 3.7V} \\[+3pt]
                \cline{2-5} \\[-6pt]
                & Avg. (mW) & Max. (mW) & Avg. (mW) & Max. (mW)  \\[+2pt]
        \midrule
        Anchor       & 171 & 699 & 129 & 545 \\
        Active Tag   & 161 & 687 & 115 & 543 \\
        Passive Tag  & 155 & 189 & 114 & 503 \\
        Custom Init. & 440 & 726 & 341 & 554 \\
        Custom Resp. & 523 & 731 & 358 & 557 \\
        \bottomrule
    \end{tabular}
\end{table}

\section{Conclusion and Future Work}\label{sec:conclusion}

We have presented a novel dataset for UWB-based localization of aerial robots. We have focused on studying the localization accuracy for ad-hoc deployments with fast self-calibration of anchor positions. Up to the authors' knowledge, the dataset presented in this paper is the largest and most complete to date. The dataset includes multiple anchor configurations, as well as data from UAVs equipped with a variable number of UWB tags. The dataset includes data from an autonomous flight with an UAV. The ground truth in all cases has been recorded using an Optitrack motion capture system. It is also the first comprehensive analysis of the UWB localization accuracy based on the UAVs speed, height and distance to the center of mass of the anchor system.
We believe that the dataset presented in this paper will enable the research community to further explore the possibilities of robust and accurate autonomous flight in GNSS-denied environments with ad-hoc localization networks via a combination of UAVs with reference ground robots.
In future work, our aim is to study further the viability of UWB for robust multi-UAV systems, as well as a distributed localization system that does not rely on anchors, but where a mesh network is built and relative positions estimated. An emphasis will also be put on fusing UWB with inertial and visual odometry data. 


\section*{Acknowledgements}

This work was supported by the Academy of Finland's AutoSOS project with grant number 328755, the Swiss National Science Foundation with grant
number 200020\_188457, and the European Union’s Horizon 2020 TERRINet project under grant agreement No 730994.

\bibliographystyle{IEEEtran}
\bibliography{main}

\end{document}